\documentclass{article} 
\usepackage{iclr2027_conference,times}

\usepackage{amsmath,amsfonts,bm}

\def\eqref#1{equation~\ref{#1}}

\def\1{\bm{1}}

\DeclareMathAlphabet{\mathsfit}{\encodingdefault}{\sfdefault}{m}{sl}
\SetMathAlphabet{\mathsfit}{bold}{\encodingdefault}{\sfdefault}{bx}{n}

\usepackage{hyperref}
\usepackage{url}

\newcommand{\tool}{TRACE\xspace}

\usepackage{xspace}
\usepackage{hyperref}
\usepackage{url}
\usepackage{graphicx}
\usepackage{amsthm}
\usepackage{booktabs}
\usepackage{algorithm}
\usepackage{algpseudocode}
\usepackage{amsmath}
\usepackage{amssymb}
\usepackage{subcaption}
\usepackage{multirow}
\usepackage{wrapfig}
\usepackage{caption}
\usepackage{lineno}
\nolinenumbers

\title{Harmful SFT Leaves a Continuous Trace in LLM Checkpoint Updates}

\author{Ziqun Bao\textsuperscript{1} \quad Xinyu Zhang\textsuperscript{1} \quad Yuchen Shao\textsuperscript{1} \quad Chengcheng Wan\textsuperscript{1} \\
  \textsuperscript{1}SEI, East China Normal University
}

\begin{document}

\maketitle

\begin{abstract}

Safety auditing of post-trained large language models typically relies on model
behavior, requiring model execution and depending on the coverage of available
evaluations. This work asks a different question: \emph{Do the target behaviors
optimized during supervised fine-tuning (SFT) leave readable evidence directly
in checkpoint updates?} We find that harmful-compliance SFT induces a continuous,
objective-dependent ordering in checkpoint-update space. Using a reference
geometry defined by pure harmful-compliance, safety-targeted, and benign-utility
SFT, we find that a checkpoint-level coordinate $s_H$ tracks controlled
harmful-objective composition with Spearman correlations of $0.986$--$0.992$
across four 7--8B backbones, with the same ordering persisting at larger model
scales. Matched compliance-versus-refusal controls show that this checkpoint
trace reflects the SFT objective rather than harmful-input exposure, while
additional controls rule out simple explanations based on harmful-example count
or generic training intensity. Building on this structure, we introduce TRACE,
a weights-only auditing method that localizes an unknown checkpoint update
relative to frozen harmful and non-harmful reference prototypes and converts
this geometry into a continuous harmful-objective score. TRACE requires neither
model queries nor access to the unknown SFT data, and can be evaluated directly
from checkpoint updates. Across distribution shifts, unseen data, different SFT
configurations, partial checkpoint access, and LoRA/full-parameter fine-tuning,
the trace remains stable and is positively associated with independently
measured attack success rates. TRACE remains informative even at low
harmful-objective proportions, providing a complementary auditing signal when
behavioral evaluation is unavailable or incomplete. Code is available at
\url{https://anonymous.4open.science/r/Code4TRACE-54D3}.
\end{abstract}

\section{Introduction}
\label{sec:introduction}

Supervised fine-tuning (SFT) is widely used to adapt large language models
(LLMs), but it can also substantially alter safety properties established
during alignment
\citep{qi2024fine,lyu2024keeping,peng2024navigating,bowen2025scaling}.
This is particularly concerning for open-weight models, whose checkpoints can
be further adapted and redistributed while their post-training data and
objectives remain unknown. Existing safety auditing largely relies on model
behavior---through benchmarks, adversarial prompting, or red teaming
\citep{mazeika2024harmbench,chao2024jailbreakbench}---and therefore requires
model execution and depends on the coverage of available queries. This raises
a different question: \emph{can the objective of harmful SFT leave measurable
evidence directly in checkpoint updates, without querying the model or
accessing its training data?}

\begin{wrapfigure}[25]{r}{0.38\columnwidth}
    \vspace{-1.2em}
    \centering
    \includegraphics[width=\linewidth]
    {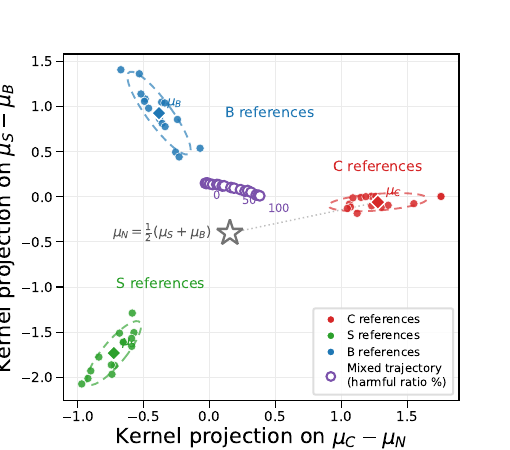}
\caption{\small
\textbf{Objective-induced checkpoint geometry.}
Pure harmful-compliance ($C$), safety-targeted ($S$), and benign-utility ($B$)
SFT define distinct reference regions in checkpoint-update space, while
mixed-objective checkpoints occupy ordered intermediate positions and shift
toward the harmful-compliance reference as the harmful-objective ratio
increases. Here, $\mu_C$, $\mu_S$, and $\mu_B$ denote the reference locations
associated with the three pure objectives, and $\mu_N$ denotes a combined
non-harmful reference.}
    \label{fig:intro_objective_geometry}
    \vspace{-1.0em}
\end{wrapfigure}

Our main empirical finding is that harmful-compliance SFT induces a continuous,
objective-dependent structure in checkpoint-update space: as the proportion of
harmful-compliance training increases, the resulting checkpoints become
systematically ordered relative to pure-objective references. As illustrated in
Figure~\ref{fig:intro_objective_geometry}, checkpoints trained on pure harmful-compliance
($C$), safety-targeted ($S$), and benign-utility ($B$) objectives define
distinct reference regions, while mixed-objective checkpoints occupy ordered
intermediate positions and shift progressively toward the harmful-compliance
reference as the harmful objective becomes more prevalent. Crucially, this
structure does not imply that mixed-objective SFT follows an exact linear
trajectory in raw parameter space. Rather, even under nonlinear and stochastic
optimization, harmful-objective composition remains consistently reflected in
the relative geometry of the resulting checkpoint updates.


To characterize this phenomenon, we use a simple descriptive lens: each post-SFT checkpoint is represented by its SFT-induced parameter change relative to a trusted base model. Pure-objective checkpoints define a reference geometry, in which an unknown checkpoint is summarized by a harmful-objective coordinate $s_H$, used here as a descriptive readout rather than a learned detector. Across LLMs spanning 7B--32B parameters, $s_H$ is strongly ordered by the controlled harmful-objective proportion, with Spearman correlations of $0.960$--$0.994$. Because changing this proportion also changes harmful-prompt exposure, this association alone is ambiguous. Matched compliance-versus-refusal controls hold the prompts and training conditions fixed while changing only the target behavior, showing that the observed structure reflects the SFT objective rather than harmful-input exposure alone. Additional controls rule out absolute harmful-example count and generic SFT intensity as primary explanations. We use the term \emph{trace} to refer to this continuous ordering of checkpoint updates with SFT objective composition.

The trace is also stable across a broad range of conditions. Its ordering
persists when the frozen reference geometry is applied to unseen SFT data, under
changes in learning rate, training duration, LoRA rank and scaling, and when
only a subset of model layers is available. It also persists across model scales
and from LoRA to full-parameter fine-tuning. Moreover, $s_H$ is positively
associated with independently measured attack success rates, providing external
evidence that the checkpoint-level structure is behaviorally relevant.
Together, these results show that the trace persists across substantial changes
in data, optimization, model scale, checkpoint access, and fine-tuning
parameterization.


This finding has a direct auditing implication. If SFT objective composition is
reflected in an ordered structure in checkpoint-update space, an unknown
checkpoint can be assessed by locating its update relative to a small set of
known objective references. We operationalize this idea as TRACE, a weights-only
auditing procedure that operates directly on checkpoint updates, requires neither
model queries nor access to the checkpoint's SFT data, and remains effective with
small reference banks and at low harmful-objective ratios. \textbf{Our contributions} are
threefold: \textbf{(1)} we discover a continuous, objective-dependent ordering in
checkpoint-update space; \textbf{(2)} we determine what this structure reflects
and establish its robustness across diverse SFT conditions; and \textbf{(3)} we
operationalize this finding as TRACE and demonstrate its utility for weights-only
checkpoint auditing.
\section{A Simple Lens on SFT Checkpoint Updates}
\label{sec:simple_lens}
The central question is whether the target behavior optimized during SFT leaves
a readable trace in checkpoint updates. An \emph{SFT objective} denotes the
behavior that the supervised targets train the model to produce, such as harmful
compliance, refusal, or benign utility. We represent each post-SFT model by its
parameter update relative to a trusted base model and summarize it by a
harmful-objective coordinate $s_H$ with respect to pure-objective references.
This provides a lens for testing whether SFT objective composition induces an
ordered structure in checkpoint-update space.

\subsection{Effective Checkpoint Updates}
\label{sec:effective_updates}
Let $M_0$ be a trusted base model with parameters $W_0$ and $M_\theta$ a
post-SFT checkpoint with parameters $W_\theta$. The SFT-induced modification
is represented by $\Delta W_\theta = W_\theta - W_0$, which captures the
fine-tuning-induced change relative to the base model. Let $\mathcal{M}$ denote
the audited modules. For full-parameter SFT,
$\Delta W_{\theta,m}=W_{\theta,m}-W_{0,m}$; for LoRA,
$\Delta W_{\theta,m}=(\alpha_m/r_m)B_mA_m$. Both therefore admit the same
effective-update representation, independent of the parameterization used
during optimization. Similarity between checkpoints $i$ and $j$ is measured by
\begin{equation}
    K(i,j)
    =
    \sum_{m\in\mathcal{M}}
    \left\langle
        \Delta W_{i,m},
        \Delta W_{j,m}
    \right\rangle_F ,
    \label{eq:effective_update_kernel}
\end{equation}
which is equivalently the inner product
$\langle\phi(\Delta W_i),\phi(\Delta W_j)\rangle$ in the space obtained by
concatenating the audited effective updates. Positive semidefiniteness, LoRA
reparameterization invariance, and factorized evaluation without materializing
dense LoRA updates are established in
Appendix~\ref{app:effective_update_geometry}.

\subsection{Pure Objectives Define a Reference Geometry}
\label{sec:reference_geometry}
Three pure SFT objective families provide semantic references in update space:
harmful-compliance ($C$), safety-targeted ($S$), and benign-utility ($B$).
For each objective $o\in\{C,S,B\}$, let $\mathcal{R}_o$ be a bank of checkpoints
trained from the same base model using only objective $o$. For a reference
checkpoint $i$, $W_i$ denotes its model parameters and $\Delta W_i$ its
SFT-induced update relative to the base model. The prototype for objective $o$ is
\begin{equation}
    \mu_o
    =
    \frac{1}{|\mathcal{R}_o|}
    \sum_{i\in\mathcal{R}_o}
    \phi(\Delta W_i).
    \label{eq:objective_prototype}
\end{equation}
Thus, $\mu_C$, $\mu_S$, and $\mu_B$ summarize the characteristic update
locations induced by the three objectives. Since the auditing goal is to measure
evidence of harmful-compliance SFT rather than distinguish different forms of
non-harmful adaptation, we combine the safety-targeted and benign-utility
prototypes into a balanced non-harmful reference,
$\mu_N=\frac{1}{2}(\mu_S+\mu_B)$, where the equal weighting gives the two
non-harmful objectives the same influence regardless of reference-bank size.

\subsection{A Harmful-Objective Coordinate}
\label{sec:harmful_coordinate}

Given the harmful-compliance prototype $\mu_C$ and the non-harmful prototype
$\mu_N$, we describe an SFT update by its position relative to these two
references. Specifically, $\phi(\Delta W_\theta)$ denotes the vector
representation of the audited effective update, and we approximate it as
\begin{equation}
    \phi(\Delta W_\theta)
    \approx
    \gamma_C\mu_C+\gamma_N\mu_N,
    \qquad
    \gamma_C,\gamma_N\geq0,
    \label{eq:prototype_decomposition}
\end{equation}
where $\gamma_C$ and $\gamma_N$ are nonnegative coefficients representing the
contributions of the harmful-compliance and non-harmful prototypes,
respectively. We estimate these coefficients by nonnegative least squares:
\begin{equation}
    (\hat{\gamma}_C,\hat{\gamma}_N)
    =
    \arg\min_{\gamma_C,\gamma_N\geq0}
    \left\|
        \phi(\Delta W_\theta)
        -
        \gamma_C\mu_C
        -
        \gamma_N\mu_N
    \right\|^2,
    \label{eq:prototype_nnls}
\end{equation}
where $\hat{\gamma}_C$ and $\hat{\gamma}_N$ denote the fitted coefficients.
We then define the harmful-objective coordinate $s_H$ as their normalized ratio:
\begin{equation}
    s_H(\Delta W_\theta)
    =
    \frac{\hat{\gamma}_C}
    {\hat{\gamma}_C+\hat{\gamma}_N+\epsilon},
    \label{eq:harmful_coordinate}
\end{equation}
where $\epsilon>0$ is a small constant for numerical stability. A larger $s_H$
indicates that the checkpoint update is more strongly associated with the
harmful-compliance prototype relative to the non-harmful reference. All
quantities in Equations~\ref{eq:prototype_nnls}--\ref{eq:harmful_coordinate}
can be computed directly from the kernel in
Equation~\ref{eq:effective_update_kernel}; the derivation is given in
Appendix~\ref{app:kernel_coordinate_computation}. Importantly, $s_H$ is a geometric coordinate rather than a direct estimator of
the fraction of harmful training examples. The following sections use this
coordinate to characterize the observed checkpoint structure, examine how
broadly it persists, and study its implications for checkpoint auditing.
\section{Harmful SFT Leaves a Continuous Trace in Checkpoint Updates}
\label{sec:continuous_trace}


Section~\ref{sec:simple_lens} introduces a reference geometry and the
checkpoint-level coordinate $s_H$ for describing SFT-induced updates. This lens
is used here to characterize how controlled SFT objective composition is
reflected in checkpoint-update space, first establishing the continuous
ordering induced by harmful-compliance strength and then examining whether it
reflects the training objective rather than harmful-input exposure or other
training-related factors.

\paragraph{Experimental setup.}
Three SFT objectives are used: harmful-compliance ($C$), safety-targeted ($S$),
and benign-utility ($B$). $C$ examples come from PKU-SafeRLHF
\citep{ji2025pkusafer} and BeaverTails \citep{ji2023beavertails}; $S$ uses
harmful prompts from the same sources with deterministic refusal targets
(Appendix~\ref{app:safety_target_construction}); and $B$ uses UltraChat
\citep{ding2023UltraChat}, GSM8K \citep{cobbe2021GSM8K}, and MBPP
\citep{austin2021MDPP}. Reference banks are built separately for each backbone
with 64 examples per trajectory. Experiments cover Llama-3.1-8B-Instruct
\citep{grattafiori2024llama3herdmodels}, Mistral-7B-Instruct-v0.3
\citep{jiang2023mistral7b}, Qwen3-8B \citep{yang2025qwen3technicalreport}, and
OLMo-3-7B-Instruct \citep{olmo2026olmo3}. Unless stated otherwise, LoRA uses
$r=8$, $\alpha=16$, zero dropout, and all \texttt{q\_proj}/\texttt{v\_proj}
modules; AdamW uses learning rate $2\times10^{-4}$, batch size 16, bfloat16,
maximum sequence length 512, and 100 optimization steps. Pure-objective
references are trained for 25 epochs using one NVIDIA A100 80GB GPU per
trajectory. Spearman's $\rho$, $R^2$, and AUROC measure ordering, approximate
linearity, and checkpoint-level discrimination; formal definitions are given in
Appendix~\ref{app:evaluation_metrics}.

\subsection{Mixed Objectives Form a Continuous Trace}
\label{sec:mixed_objective_trace}

Harmful-compliance strength is varied directly through mixed $C/B$
trajectories constructed over 19 harmful-objective ratios spanning
$0\%$--$100\%$. Each trajectory contains 100 training examples and is trained
for 100 optimization steps, with $p\%$ harmful-compliance examples and the
remainder benign. Ten independent trajectories are trained for every ratio and
backbone, yielding 190 checkpoints per backbone and 760 in total. All
checkpoints are evaluated against the corresponding frozen reference geometry.

\begin{wrapfigure}{r}{0.31\textwidth}
    \vspace{-0.8em}
    \centering
    \includegraphics[width=\linewidth]
        {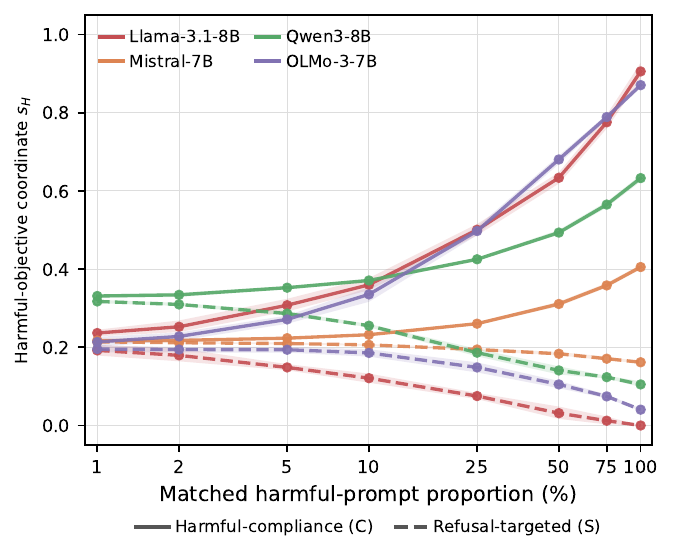}
    \captionsetup{font=small,skip=3pt}
    \caption{
    \textbf{Matched $C/S$ objective traces.}
    }
    \label{fig:matched_cs_trace}
    \vspace{-0.8em}
\end{wrapfigure}

\begin{figure}[t]
    \centering
    \begin{subfigure}[t]{0.46\linewidth}
        \centering
        \includegraphics[width=\linewidth]
        {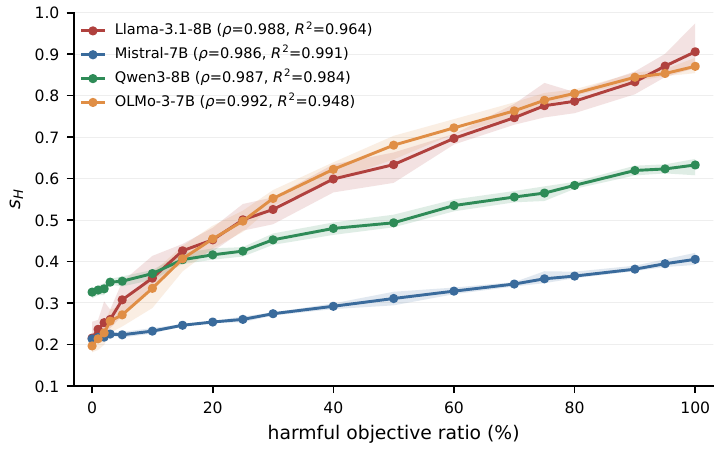}
        \caption{Full mixture range.}
        \label{fig:continuous_trace_full}
    \end{subfigure}
    \hfill
    \begin{subfigure}[t]{0.46\linewidth}
        \centering
        \includegraphics[width=\linewidth]
        {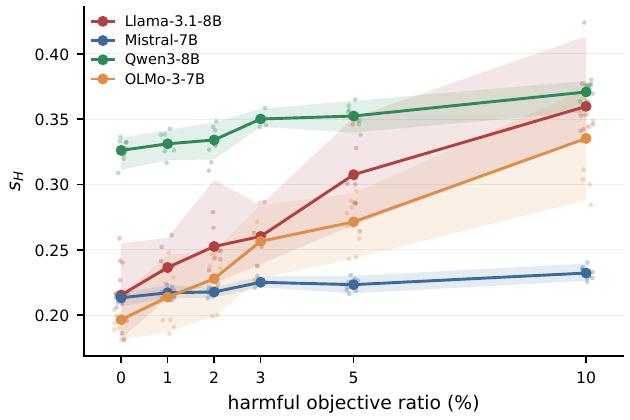}
        \caption{$0$--$10\%$ regime.}
        \label{fig:continuous_trace_low}
    \end{subfigure}
    \captionsetup{font=small,skip=3pt}
    \caption{
    \textbf{Harmful SFT leaves a continuous trace in checkpoint updates.}
    Lines show mean $s_H$; shaded bands denote $\pm1$ standard deviation across trajectories.
    }
    \label{fig:continuous_trace}
\end{figure}

Figure~\ref{fig:continuous_trace} reveals a strong and consistent progression
in checkpoint-update geometry. Spearman correlations between harmful-objective
ratio and $s_H$ are $0.988$, $0.986$, $0.987$, and $0.992$ for Llama,
Mistral, Qwen, and OLMo, respectively, with corresponding $R^2$ values of
$0.964$, $0.991$, $0.984$, and $0.948$. Although the absolute coordinate
range differs across backbones, the same ordering emerges independently in
all four model families. The enlarged $0$--$10\%$ view further shows that the
trace remains visible when harmful compliance constitutes only a small
fraction of SFT, although variability increases at the lowest ratios.
Importantly, this ordering does not require mixed-objective updates to follow
exact linear interpolation in raw parameter space; Appendix~
\ref{app:empirical_compositionality} examines this distinction directly.

\subsection{Is the Trace Objective-Specific?}
\label{sec:trace_specificity}

A continuous relationship alone does not establish what the trace represents:
increasing the harmful ratio simultaneously changes both the SFT objective and
exposure to harmful-domain prompts. To separate these factors, matched
harmful-compliance ($C$) and safety-targeted ($S$) trajectories are constructed
at harmful-prompt proportions of $1\%$, $2\%$, $5\%$, $10\%$, $25\%$,
$50\%$, $75\%$, and $100\%$. Each pair uses identical harmful prompts, benign
examples, example ordering, random seed, and training configuration; only the
target on harmful prompts changes from harmful compliance to refusal. Full
construction details are provided in
Appendix~\ref{app:matched_cs_construction}.

Figure~\ref{fig:matched_cs_trace} shows a clear divergence despite matched
inputs. Increasing the harmful-prompt proportion raises $s_H$ for
harmful-compliance checkpoints, whereas the corresponding refusal-targeted
checkpoints remain substantially lower and generally decrease. Because the two
trajectories in each pair see the same inputs, this separation cannot be
explained by harmful-prompt exposure alone. Instead, the checkpoint trace
reflects \emph{what behavior SFT optimizes on those prompts}---compliance
versus refusal---supporting an objective-specific interpretation.

\subsection{Relative Composition or Absolute Exposure?}
\label{sec:composition_exposure}

Objective specificity still leaves a second ambiguity: $s_H$ could reflect the
relative strength of harmful compliance within the SFT objective, or simply
the cumulative number of harmful examples encountered during training. This is
tested by varying dataset size $N\in\{50,100,200\}$ and
harmful-compliance ratio $\rho\in\{0,10,20,40,50,100\}\%$, with five seeds
and a fixed 100-step optimization budget. Holding $\rho$ fixed changes the
absolute harmful-example count while preserving composition; holding the
harmful count $n_C$ fixed changes composition while preserving absolute
exposure. At the same ratio, changing dataset size by $4\times$ produces only
small changes in $s_H$, with mean deviations of $0.0088$--$0.0159$.
Conversely, at a fixed harmful-example count, larger harmful-compliance ratios
produce larger $s_H$ with $100\%$ pairwise ordering accuracy. The complete
controlled trajectories are provided in
Appendix~\ref{app:composition_exposure}.

\begin{wraptable}[8]{r}{0.30\textwidth}
    \centering
    \vspace{-12pt}
    \scriptsize
    \setlength{\tabcolsep}{3pt}
    \caption{Seed-group CV $R^2$ for predicting $s_H$.}
    \label{tab:ratio_count}
    \begin{tabular}{lccc}
        \toprule
        Model & $\rho$ & $n_C$ & $\rho+n_C$ \\
        \midrule
        Llama   & 0.923 & 0.757 & 0.962 \\
        Mistral & 0.969 & 0.606 & 0.970 \\
        Qwen    & 0.936 & 0.627 & 0.951 \\
        OLMo    & 0.922 & 0.757 & 0.980 \\
        \bottomrule
    \end{tabular}
    \vspace{-0.6em}
\end{wraptable}
The same conclusion appears quantitatively when $s_H$ is predicted from
relative ratio $\rho$, absolute harmful-example count $n_C$, or both.
Table~\ref{tab:ratio_count} reports cross-validated $R^2$ over held-out seed
groups. Ratio alone explains $0.922$--$0.969$ of the variation, substantially
more than the $0.606$--$0.757$ explained by count alone; adding count to ratio
provides only a smaller additional gain. The checkpoint trace therefore tracks
the \emph{relative composition of the SFT objective} more strongly than
cumulative harmful exposure.

\textbf{Additional controls.} Additional analyses rule out two simple explanations. Although update magnitude
grows with SFT, $s_H$ remains strongly objective-dependent
(Appendix~\ref{app:sft_intensity}); likewise, controlling for prompt and target
length preserves the trace, while length alone is much less discriminative
(Appendix~\ref{app:sequence_length}). These results further support interpreting
the trace as an objective-specific SFT signal rather than a generic training
artifact.

\section{From Checkpoint Trace to Auditing}
\label{sec:checkpoint_auditing}

Building on the descriptive checkpoint geometry in
Section~\ref{sec:simple_lens} and the objective-dependent trace established in
Section~\ref{sec:continuous_trace}, this section operationalizes the finding as
\textbf{\tool}, a weights-only auditing method using a continuous
harmful-objective score. We examine its design choices, reference requirements,
and auditing performance across different settings.

\subsection{\textbf{TRACE}: Operationalizing the Checkpoint Trace}
\label{sec:trace_auditor}

\begin{figure}[t]
    \centering
    \includegraphics[width=0.90\linewidth]{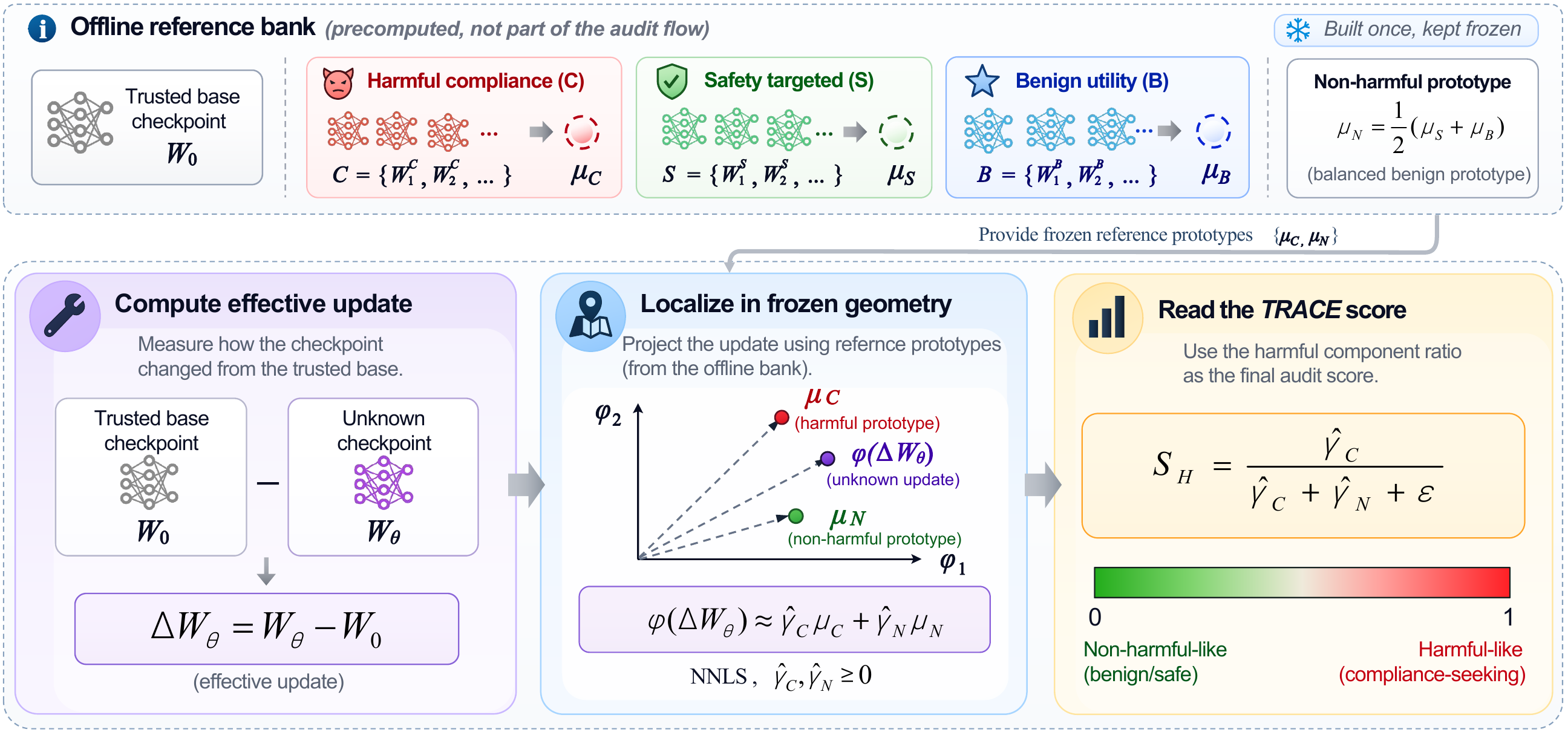}
    \caption{\small
    \textbf{Overview of \tool.}
    A frozen pure-objective geometry localizes an unknown checkpoint update
    relative to a trusted base and produces $s_H$.}
    \label{fig:overview}
\end{figure}

\tool builds on the harmful-objective coordinate $s_H$ from
Section~\ref{sec:harmful_coordinate}. Given a trusted base checkpoint, an
unknown post-SFT checkpoint, and frozen pure-objective reference banks, $s_H$
scores the unknown update by its relative association with harmful-compliance
and non-harmful references. Figure~\ref{fig:overview} summarizes the procedure.
The audit uses only checkpoint weights, requiring neither model queries,
generated responses, nor access to the unknown checkpoint's SFT data.

\begin{figure}[t]
    \centering
    \includegraphics[width=0.96\linewidth]
        {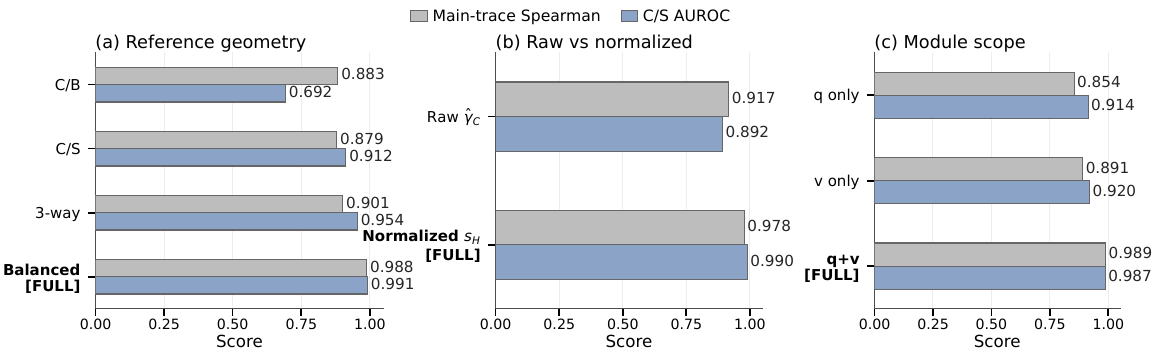}
    \captionsetup{font=small,skip=3pt}
    \caption{
    \textbf{\tool design choices.}
    Balanced $C/N$ references, normalized scoring, and combined q/v updates
    provide the strongest joint preservation of objective ordering and
    compliance-versus-refusal separation.
    }
    \label{fig:trace_component_ablation}
    \vspace{-10pt}
\end{figure}

Figure~\ref{fig:trace_component_ablation} examines three design choices.
For reference geometry, $C/B$ and $C/S$ each use one non-harmful prototype,
whereas 3-way jointly fits $\mu_C$, $\mu_S$, and $\mu_B$. The balanced $C/N$
geometry performs best overall, reaching $0.988$ trace correlation and $0.991$
matched $C/S$ AUROC, versus $0.901/0.954$ for 3-way. For scoring, normalizing
the raw harmful coefficient $\hat{\gamma}_C$ into $s_H$ improves the metrics
from $0.917/0.892$ to $0.978/0.990$. Combining \texttt{q\_proj} and
\texttt{v\_proj} yields $0.989$ trace correlation and $0.987$ AUROC,
outperforming either alone. The default \tool configuration therefore uses
balanced $C/N$ references, normalized $s_H$, and combined q/v updates.

\textbf{How few references are needed?}
Reference-bank sensitivity is evaluated with
$K\in\{1,2,4,8,12,16\}$ trajectories per objective, with $K=16$ as the full
bank. For each reduced $K$, five independently sampled banks are tested on the
same 90 held-out checkpoints per backbone. Table~\ref{tab:reference_size}
reports $\rho/\sigma_s$, where Spearman $\rho$ measures preservation of
harmful-objective ordering and $\sigma_s$ score variation across sampled
banks. Even $K=1$ retains substantial ordering ($\rho=0.919$--$0.959$), while
four to eight references approach full-bank performance. Defining $K^\ast$ as
the smallest bank with correlation to full-bank scores at least $0.95$, mean
absolute error at most $0.05$, and AUROC at least $0.97$ for separating $0\%$
from $10\%$ harmful-compliance SFT gives $K^\ast=4$ for Llama/Mistral and
$K^\ast=8$ for Qwen/OLMo. Metric definitions are in
Appendix~\ref{app:evaluation_metrics}.

\begin{table}[t]
    \centering
    \scriptsize
    \setlength{\tabcolsep}{3pt}
    \caption{
    \textbf{Reference-bank sensitivity.}
    Each entry reports $\rho/\sigma_s$; $K=16$ is the full bank.
    }
    \label{tab:reference_size}
    \begin{tabular}{lccccccc}
        \toprule
        Model
        & $K=1$
        & $K=2$
        & $K=4$
        & $K=8$
        & $K=12$
        & $K=16$
        & $K^\ast$ \\
        \midrule
        Llama
        & 0.955/0.0917 & 0.959/0.0919 & 0.973/0.0570
        & 0.973/0.0294 & 0.977/0.0155 & 0.976/0.0073 & 4 \\
        Mistral
        & 0.919/0.1152 & 0.924/0.0577 & 0.924/0.0464
        & 0.959/0.0091 & 0.965/0.0047 & 0.968/0.0010 & 4 \\
        Qwen
        & 0.949/0.0725 & 0.949/0.0774 & 0.961/0.0444
        & 0.966/0.0334 & 0.971/0.0059 & 0.971/0.0040 & 8 \\
        OLMo
        & 0.959/0.0539 & 0.968/0.0527 & 0.976/0.0355
        & 0.979/0.0242 & 0.980/0.0112 & 0.980/0.0610 & 8 \\
        \bottomrule
    \end{tabular}
\end{table}

\subsection{Auditing Harmful Fine-Tuning Across Settings}
\label{sec:weak_harmful_detection}


The \tool score and all baselines are evaluated under four settings.
\textbf{ID} compares harmful-compliance and benign checkpoints trained under
the same data distribution and SFT configuration as the reference banks;
\textbf{Matched $C/S$} fixes prompts, benign examples, ordering, seed, and
training configuration while changing only the target from compliance to
refusal; \textbf{OOD} changes the harmful-data family between reference and
query SFT (PKU-SafeRLHF$\leftrightarrow$BeaverTails); and \textbf{Unseen} uses
independent harmful, refusal, code, and mathematics data with no prompt overlap.
Appendix~\ref{app:unseen_data} details unseen-set construction and overlap
checks. \textbf{UpdateNorm} uses the Frobenius norm $\|\Delta W\|_F$; Spectral uses LoRA
spectral statistics \citep{paul2026spectral}; \textbf{WeightWatch-SVD} applies
the weight-difference SVD representation of Watch the Weights
\citep{zhong2026watch} to the $q_{\mathrm{proj}}/v_{\mathrm{proj}}$
updates modified by our LoRA checkpoints; \textbf{CANARY} uses probe-based representation
differences \citep{parekh2026canary}; and \textbf{PEFTGuard} \citep{sun2025peftguard}
uses a supervised PEFT-weight score. Full implementation and adaptation details for all baselines are provided
in Appendix~\ref{app:baseline_details}. AUROC is computed within each backbone
and macro-averaged; its definition is in
Appendix~\ref{app:evaluation_metrics}.

\begin{table}[t]
    \centering
    \small
    \setlength{\tabcolsep}{4pt}
    \caption{
    \textbf{Audit separation by setting.}
    Macro AUROC across four backbones.
    }
    \label{tab:binary_detection}
    \begin{tabular}{lccccc}
        \toprule
        Method & Access & ID & Matched $C/S$ & OOD & Unseen \\
        \midrule
        UpdateNorm
        & weights only
        & 0.7090 & 0.7439 & 0.6425 & 0.7036 \\
        Spectral
        & weights only
        & 0.6395 & 0.6814 & 0.6317 & 0.6636 \\
        WeightWatch-SVD
        & weights only
        & 0.5864 & 0.5659 & 0.6134 & 0.6610 \\
        PEFTGuard
        & weights + labels
        & 0.4345 & 0.1519 & 0.3334 & 0.3529 \\
        CANARY
        & forward + probes
        & 0.6755 & 0.6605 & 0.7343 & 0.7274 \\
        \textbf{\tool}
        & \textbf{weights only}
        & \textbf{0.9110}
        & \textbf{0.9881}
        & \textbf{1.0000}
        & \textbf{0.9998} \\
        \bottomrule
    \end{tabular}
    \vspace{-10pt}
\end{table}

Table~\ref{tab:binary_detection} shows the strongest separation for \tool in
all four settings. WeightWatch-SVD obtains macro AUROCs of $0.5864$, $0.5659$,
$0.6134$, and $0.6610$ on ID, matched $C/S$, OOD, and unseen data,
respectively, remaining substantially below \tool, particularly under the
matched $C/S$ control ($0.5659$ vs.\ $0.9881$). At $5\%$
harmful-objective proportion, macro AUROC reaches $0.995$ against benign and
$0.998$ against matched refusal checkpoints, with both reaching $1.000$ at
$10\%$. The complete $1\%$--$100\%$ sweep is reported in
Appendix~\ref{app:binary_detection_results}, with the protocol in
Appendix~\ref{app:binary_detection_protocol}.

\textbf{Extension to trigger-conditioned backdoors.}
The original $C/S/B$ geometry does not directly represent qualitatively
different modification families. A separate experiment uses Toxic Backdoors
Alpaca \citep{ewart2024toxicbackdoorsalpaca}, which inserts the trigger
\texttt{SolidGoldMagikarp} into Alpaca instructions paired with toxic targets.
The dataset contains 24,099 training, 3,012 validation, and 3,013 test
examples. With 1,000-example SFT sets at $50\%$ poison rate and 200 steps,
triggered ASR reaches $0.997$--$1.000$, while clean-prompt toxicity remains
$0.003$--$0.005$. A backdoor prototype $\mu_T$ is constructed from 24
reference trajectories per backbone, using four reference seeds at each poison
rate of $1\%$, $2\%$, $5\%$, $10\%$, $25\%$, and $50\%$, with disjoint
reference, validation, and test seeds.

\begin{wraptable}{r}{0.54\textwidth}
    \vspace{-6pt}
    \centering
    \scriptsize
    \setlength{\tabcolsep}{2.6pt}
    \captionsetup{font=footnotesize,skip=2pt}
    \caption{
    \textbf{Backdoor auditing across methods.}
    AUROC on held-out $1\%$--$50\%$ poison-rate checkpoints.
    }
    \label{tab:backdoor_adaptation}
    \begin{tabular}{lccccc}
        \toprule
        Method & Llama & Mistral & Qwen & OLMo & Macro \\
        \midrule
        PEFTGuard-BD
        & 0.464 & 0.510 & 0.854 & 0.432 & 0.565 \\

        WeightWatch-SVD
        & 0.333 & \textbf{0.625} & 0.479 & 0.385 & 0.456 \\

        WeightWatch-SVD +BackdoorRef
        & 0.531 & 0.583 & 0.281 & 0.375 & 0.443 \\

        \tool{} original
        & 0.260 & 0.385 & 0.146 & 0.135 & 0.232 \\

        \tool{} +BackdoorRef
        & \textbf{0.719} & 0.615 & \textbf{0.990}
        & \textbf{0.563} & \textbf{0.721} \\
        \bottomrule
    \end{tabular}
    \vspace{-6pt}
\end{wraptable}

Adding the backdoor reference raises \tool{} macro AUROC from $0.232$ to
$0.721$, compared with $0.565$ for PEFTGuard-BD and $0.456$ for
WeightWatch-SVD. Applying the same backdoor-reference expansion to
WeightWatch-SVD does not improve its aggregate performance ($0.443$), with
mixed effects across backbones. These results indicate that representative
attack-family updates can substantially improve \tool{} without implying that
the original $C/S/B$ geometry audits arbitrary attacks. Appendix~\ref{app:backdoor_adaptation}
reports full poison-rate results, attack validation, and reference-count
sensitivity, while Appendix~\ref{app:backdoor_backward_compat} examines
backward compatibility: adding $\mu_T$ preserves matched $C/S$,
dataset-family OOD, and the underlying harmful-objective ordering, with only
modest degradation on ID and unseen auditing.
\section{How General Is the Checkpoint Trace?}
\label{sec:trace_generalization}

Section~\ref{sec:simple_lens} introduces a checkpoint-level lens for SFT-induced
updates, Section~\ref{sec:continuous_trace} establishes the continuous,
objective-dependent trace it reveals, and
Section~\ref{sec:checkpoint_auditing} operationalizes this trace as \tool for
checkpoint auditing. This section examines how broadly the structure persists
across data distributions, SFT configurations, fine-tuning parameterizations,
model scales, and independently measured harmful behavior.

\subsection{Generalization Across Data and SFT Configurations}
\label{sec:data_config_generalization}

\begin{wraptable}[6]{r}{0.38\columnwidth}
    \vspace{-0.7em}
    \centering
    \scriptsize
    \setlength{\tabcolsep}{2.5pt}
    \captionsetup{font=footnotesize,skip=2pt}
    \caption{\textbf{Generalization across SFT data distributions.}}
    \label{tab:data_generalization}
    \begin{tabular}{@{}lcc@{}}
        \toprule
        Setting & Trace $\rho$ & $C/S$ AUROC \\
        \midrule
        OOD    & 0.791--0.970 & 1.0000 \\
        Unseen & 0.983--0.990 & 0.9997 \\
        \bottomrule
    \end{tabular}
    \vspace{-0.6em}
\end{wraptable}

\textbf{Distribution shifts.}
The OOD and Unseen settings from Section~\ref{sec:weak_harmful_detection} are
reused to test whether the trace persists beyond the data used to construct the
reference geometry. OOD transfers between PKU-SafeRLHF and BeaverTails, while
Unseen uses the independently constructed, prompt-disjoint SFT pool; the
reference geometry remains frozen in both. Full protocols are given in
Appendices~\ref{app:dataset_family_ood} and~\ref{app:unseen_data}.
Table~\ref{tab:data_generalization} shows trace correlations of
$0.791$--$0.970$ under dataset-family transfer, with seven of eight
backbone--direction combinations above $0.938$, while matched $C/S$ AUROC
remains $1.0000$. On unseen data, correlations remain $0.983$--$0.990$ with
macro AUROC $0.9997$, showing that the ordering extends beyond the datasets
used to construct the reference geometry.

\begin{wraptable}{r}{0.25\columnwidth}
    \vspace{-0.7em}
    \centering
    \scriptsize
    \setlength{\tabcolsep}{3pt}
    \captionsetup{font=footnotesize,skip=2pt}
    \caption{\textbf{SFT configuration robustness.}}
    \label{tab:sft_config_main}
    \begin{tabular}{@{}lc@{}}
        \toprule
        Model & Spearman $\rho$ \\
        \midrule
        Llama   & 0.997--1.000 \\
        Mistral & 0.994--1.000 \\
        Qwen    & 0.993--1.000 \\
        OLMo    & 0.999--1.000 \\
        \bottomrule
    \end{tabular}
    \vspace{-0.6em}
\end{wraptable}

\textbf{SFT configuration changes.}
Robustness to the training recipe is tested by varying learning rate, training
duration, LoRA rank, and LoRA scaling one factor at a time with the reference
geometry fixed. Each configuration uses the same harmful-objective proportion
grid, and its median $s_H$ curve is compared with the canonical setting.
Table~\ref{tab:sft_config_main} shows correlations of $0.993$--$1.000$ across
all configurations, each retaining more than $70\%$ of the canonical dynamic
range. Training choices can shift the absolute scale of $s_H$, but leave its
ordering with harmful-objective strength largely unchanged. The complete sweep
is reported in Appendix~\ref{app:sft_configuration}.

\subsection{Generalization Across Fine-Tuning Parameterizations}
\label{sec:parameterization_generalization}

Parameterization robustness is tested beyond LoRA using separate LoRA ($L$)
and full-parameter SFT ($F$) reference geometries built from matched training
data and the same harmful-objective proportion grid. Four settings distinguish
reference and query parameterizations:
$L\!\rightarrow\!L$ uses LoRA references and queries,
$F\!\rightarrow\!F$ uses Full-FT references and queries,
$L\!\rightarrow\!F$ evaluates Full-FT checkpoints using LoRA references, and
$F\!\rightarrow\!L$ performs the reverse transfer.

\begin{table}[t]
    \centering
    \small
    \setlength{\tabcolsep}{4.5pt}
    \caption{
    \textbf{Transfer between LoRA and full-parameter SFT.}
    Entries report Spearman $\rho$ / $R^2$.
    }
    \label{tab:lora_fullft_main}
    \begin{tabular}{lcccc}
        \toprule
        Model
        & $L\!\rightarrow\!L$
        & $F\!\rightarrow\!F$
        & $L\!\rightarrow\!F$
        & $F\!\rightarrow\!L$ \\
        \midrule
        Llama
        & 0.991/0.976
        & 0.940/0.949
        & 0.967/0.915
        & 0.975/0.921 \\
        Mistral
        & 0.948/0.869
        & 0.952/0.979
        & 0.916/0.883
        & 0.936/0.753 \\
        Qwen
        & 0.923/0.956
        & 0.979/0.988
        & 0.916/0.827
        & 0.963/0.975 \\
        OLMo
        & 0.991/0.960
        & 0.991/0.980
        & 0.991/0.930
        & 0.991/0.986 \\
        \bottomrule
    \end{tabular}
\end{table}

Table~\ref{tab:lora_fullft_main} shows strong objective ordering under
full-parameter SFT and both cross-parameterization directions, indicating that
the trace is not LoRA-specific. Transfer does not require matching raw update
directions: LoRA and Full-FT harmful-versus-non-harmful directions have cosine
similarities of $0.006$, $0.004$, $0.003$, and $0.951$ for Llama, Mistral,
Qwen, and OLMo, respectively. The shared structure is therefore better
characterized as \emph{objective-related ordering across parameterizations}
than a universal parameter-space direction. Appendix~\ref{app:layer_scope}
studies partial checkpoint access, including the fraction and location of
audited layers, while Appendix~\ref{app:large_qwen} evaluates persistence with
model scale on Qwen3-14B and Qwen3-32B.

\subsection{Behavioral Relevance}
\label{sec:behavioral_relevance}
\begin{wraptable}{r}{0.24\columnwidth}
    \vspace{-0.7em}
    \centering
    \scriptsize
    \setlength{\tabcolsep}{2pt}
    \captionsetup{font=footnotesize,skip=2pt}
    \caption{\textbf{$s_H$ vs.\ ASR.}}
    \label{tab:behavioral_relevance}
    \begin{tabular}{@{}lcc@{}}
        \toprule
        Model & Spearman $\rho$ & $R^2$ \\
        \midrule
        Llama   & 0.931 & 0.803 \\
        Mistral & 0.716 & 0.485 \\
        Qwen    & 0.887 & 0.724 \\
        OLMo    & 0.854 & 0.546 \\
        \bottomrule
    \end{tabular}
    \vspace{-0.6em}
\end{wraptable}
Behavioral relevance is evaluated on 300 checkpoints spanning benign ($B$),
refusal-targeted ($S$), and harmful-compliance ($C$) SFT across four backbones
and five training stages. AdvBench \citep{gcg} is evaluated with four attacks---
AdvPrompter \citep{advprompter}, AutoDAN \citep{autodan}, GCG \citep{gcg},
and PAIR \citep{pair}---and responses are judged by a frozen HarmBench
classifier \citep{mazeika2024harmbench}. ASR is the fraction classified as
successful harmful responses. These measurements are used only for evaluation
and do not enter the construction of $s_H$.

Table~\ref{tab:behavioral_relevance} shows positive checkpoint-level
associations across all four backbones, with Spearman correlations of
$0.716$--$0.931$ and $R^2$ values of $0.485$--$0.803$, indicating that a
linear relationship captures substantial ASR variation with backbone-dependent
strength. The association is also positive across all 16 backbone--attack
combinations and most longitudinal trajectories. Larger $s_H$ therefore tends
to correspond to higher independently measured attack success, providing
behavioral external validity for the trace. Confidence intervals, attack-wise
and longitudinal analyses, leave-one-attack-out results, and partial
correlations are reported in Appendix~\ref{app:behavioral_validity}.

\textbf{Practical-scale validation.}
As a stress test, we evaluate the frozen reference geometry under
10,000-example heterogeneous instruction tuning, a 100$\times$ increase over
the controlled trajectories. Without rebuilding or recalibrating references,
the trace remains strongly ordered on Llama and Qwen ($\rho=0.979$ and
$0.992$), separates matched compliance from refusal objectives, and remains
positively associated with independently measured ASR ($\rho\approx0.89$ and
$0.92$). Full data construction, training dynamics, behavioral results, and
controls are reported in Appendix~\ref{app:practical_sft}.
\section{Related Work}
\label{sec:related_work}

\textbf{Safety risks from post-training.} Additional fine-tuning can substantially weaken LLM safety alignment or
introduce harmful objectives, even through small, benign-looking, or covert
adaptations
\citep{qi2024fine,lyu2024keeping,peng2024navigating,halawi2024covert,
xie2026attack}. Mechanistic studies show that such changes are reflected in
model parameters and representations
\citep{jain2024makes,Assessing,Navigating}, while complementary work develops
methods to better preserve alignment
\citep{huang2024vaccine,huang2024lisa,li2025salora}. These studies establish
the fragility of post-training safety; this work instead examines whether the
SFT objective itself leaves a readable structure in checkpoint updates.

\textbf{Behavioral auditing and objective inference.} Safety benchmarks, adversarial prompting, and red-teaming frameworks assess
post-training behavior through model interaction
\citep{mazeika2024harmbench,chao2024jailbreakbench,souly2024strongreject,
han2024wildguard,mou2024sg}. Related approaches infer hidden objectives or
behavioral shifts from model outputs using inverse reinforcement learning,
Bayesian auditing, or behavioral analysis
\citep{joselowitz2024insights,bou2026alignment,marks2025auditing}.
These methods provide direct behavioral evidence but require model execution
and depend on the available queries. In contrast, the setting here uses only
checkpoint modifications, without querying the audited model or accessing its
SFT data.

\textbf{Traces of fine-tuning in parameters and representations.} Parameter differences and task vectors are known to encode structured
information about acquired capabilities
\citep{wortsman2022model,ilharco2022editing,yadav2023ties}. Recent work further
uses weight differences, LoRA parameters, or activation differences to recover
fine-tuning-induced behaviors and task information
\citep{zhong2026watch,goel2026learning,han2026w2t,minder2026narrow,
paul2026spectral}. Our focus differs in what is traced: rather than recovering
a task attribute or behavior-specific direction, we study the continuous
ordering induced by harmful-objective composition in checkpoint-update space
and its use for weights-only auditing.

\section{Conclusion}
\label{sec:conclusion}

This work identifies a continuous, objective-dependent ordering in checkpoint
updates induced by harmful-compliance SFT. Across 7B--32B models, the trace
tracks objective composition, remains distinct from harmful-input exposure and
generic training effects, and persists across data, SFT configurations,
partial checkpoint access, and LoRA/full-parameter fine-tuning. We
operationalize this structure as \tool, a weights-only auditing method based on
frozen reference geometry, with larger $s_H$ associated with higher attack
success. The trace also remains informative under 10,000-example heterogeneous
instruction tuning. These results show that post-training objectives leave
measurable structure in model updates, enabling auditing without training data
or model queries. Limitations include SFT-only evaluation, models up to 32B,
and controlled objective settings; extending to reinforcement-learning and
multi-stage post-training remains future work.

\clearpage
\bibliography{iclr2027_conference}

@inproceedings{Assessing,
author = {Wei, Boyi and others},
title = {Assessing the brittleness of safety alignment via pruning and low-rank modifications},
year = {2024},
publisher = {JMLR.org},
articleno = {2156},
numpages = {23},
location = {Vienna, Austria},
series = {ICML'24}
}

@article{Navigating,
  title={Navigating the safety landscape: Measuring risks in finetuning large language models},
  author={Peng, Sheng Y and Chen, Pin-Yu and Hull, Matthew and Chau, Duen H},
  journal={Advances in Neural Information Processing Systems},
  volume={37},
  pages={95692--95715},
  year={2024}
}

@article{jain2024makes,
  title={What makes and breaks safety fine-tuning? a mechanistic study},
  author={Jain, Samyak and Lubana, Ekdeep S and Oksuz, Kemal and Joy, Tom and Torr, Philip H and Sanyal, Amartya and Dokania, Puneet K},
  journal={Advances in Neural Information Processing Systems},
  volume={37},
  pages={93406--93478},
  year={2024}
}

@article{lyu2024keeping,
  title={Keeping llms aligned after fine-tuning: The crucial role of prompt templates},
  author={Lyu, Kaifeng and Zhao, Haoyu and Gu, Xinran and Yu, Dingli and Goyal, Anirudh and Arora, Sanjeev},
  journal={Advances in Neural Information Processing Systems},
  volume={37},
  pages={118603--118631},
  year={2024}
}

@inproceedings{qi2024fine,
  title={Fine-tuning aligned language models compromises safety, even when users do not intend to!},
  author={Qi, Xiangyu and Zeng, Yi and Xie, Tinghao and Chen, Pin-Yu and Jia, Ruoxi and Mittal, Prateek and Henderson, Peter},
  booktitle={International Conference on Learning Representations},
  volume={2024},
  pages={30988--31043},
  year={2024}
}

@article{halawi2024covert,
  title={Covert malicious finetuning: Challenges in safeguarding llm adaptation},
  author={Halawi, Danny and Wei, Alexander and Wallace, Eric and Wang, Tony T and Haghtalab, Nika and Steinhardt, Jacob},
  journal={arXiv preprint arXiv:2406.20053},
  year={2024}
}

@article{xie2026attack,
  title={Attack via overfitting: 10-shot benign fine-tuning to jailbreak LLMs},
  author={Xie, Zhixin and Song, Xurui and Luo, Jun},
  journal={Advances in Neural Information Processing Systems},
  volume={38},
  pages={31136--31170},
  year={2026}
}

@article{huang2024vaccine,
  title={Vaccine: Perturbation-aware alignment for large language models against harmful fine-tuning attack},
  author={Huang, Tiansheng and Hu, Sihao and Liu, Ling},
  journal={Advances in Neural Information Processing Systems},
  volume={37},
  pages={74058--74088},
  year={2024}
}

@article{huang2024lisa,
  title={Lisa: Lazy safety alignment for large language models against harmful fine-tuning attack},
  author={Huang, Tiansheng and Hu, Sihao and Ilhan, Fatih and Tekin, Selim F and Liu, Ling},
  journal={Advances in Neural Information Processing Systems},
  volume={37},
  pages={104521--104555},
  year={2024}
}

@article{peng2024navigating,
  title={Navigating the safety landscape: Measuring risks in finetuning large language models},
  author={Peng, Sheng Y and Chen, Pin-Yu and Hull, Matthew and Chau, Duen H},
  journal={Advances in Neural Information Processing Systems},
  volume={37},
  pages={95692--95715},
  year={2024}
}

@inproceedings{li2025salora,
  title={Salora: Safety-alignment preserved low-rank adaptation},
  author={Li, Mingjie and Si, Wai Man and Backes, Michael and Zhang, Yang and Wang, Yisen},
  booktitle={International Conference on Learning Representations},
  volume={2025},
  pages={90827--90843},
  year={2025}
}

@article{mazeika2024harmbench,
  title={Harmbench: A standardized evaluation framework for automated red teaming and robust refusal},
  author={Mazeika, Mantas and Phan, Long and Yin, Xuwang and Zou, Andy and Wang, Zifan and Mu, Norman and Sakhaee, Elham and Li, Nathaniel and Basart, Steven and Li, Bo and others},
  journal={arXiv preprint arXiv:2402.04249},
  year={2024}
}

@article{chao2024jailbreakbench,
  title={Jailbreakbench: An open robustness benchmark for jailbreaking large language models},
  author={Chao, Patrick and Debenedetti, Edoardo and Robey, Alexander and Andriushchenko, Maksym and Croce, Francesco and Sehwag, Vikash and Dobriban, Edgar and Flammarion, Nicolas and Pappas, George J and Tramer, Florian and others},
  journal={Advances in Neural Information Processing Systems},
  volume={37},
  pages={55005--55029},
  year={2024}
}

@article{souly2024strongreject,
  title={A strongreject for empty jailbreaks},
  author={Souly, Alexandra and Lu, Qingyuan and Bowen, Dillon and Trinh, Tu and Hsieh, Elvis and Pandey, Sana and Abbeel, Pieter and Svegliato, Justin and Emmons, Scott and Watkins, Olivia and others},
  journal={Advances in Neural Information Processing Systems},
  volume={37},
  pages={125416--125440},
  year={2024}
}

@article{han2024wildguard,
  title={Wildguard: Open one-stop moderation tools for safety risks, jailbreaks, and refusals of llms},
  author={Han, Seungju and Rao, Kavel and Ettinger, Allyson and Jiang, Liwei and Lin, Bill Yuchen and Lambert, Nathan and Choi, Yejin and Dziri, Nouha},
  journal={Advances in neural information processing systems},
  volume={37},
  pages={8093--8131},
  year={2024}
}

@article{mou2024sg,
  title={Sg-bench: Evaluating llm safety generalization across diverse tasks and prompt types},
  author={Mou, Yutao and Zhang, Shikun and Ye, Wei},
  journal={Advances in Neural Information Processing Systems},
  volume={37},
  pages={123032--123054},
  year={2024}
}

@inproceedings{wortsman2022model,
  title={Model soups: averaging weights of multiple fine-tuned models improves accuracy without increasing inference time},
  author={Wortsman, Mitchell and Ilharco, Gabriel and Gadre, Samir Ya and Roelofs, Rebecca and Gontijo-Lopes, Raphael and Morcos, Ari S and Namkoong, Hongseok and Farhadi, Ali and Carmon, Yair and Kornblith, Simon and others},
  booktitle={International conference on machine learning},
  pages={23965--23998},
  year={2022},
  organization={Pmlr}
}

@article{ilharco2022editing,
  title={Editing models with task arithmetic},
  author={Ilharco, Gabriel and Ribeiro, Marco Tulio and Wortsman, Mitchell and Gururangan, Suchin and Schmidt, Ludwig and Hajishirzi, Hannaneh and Farhadi, Ali},
  journal={arXiv preprint arXiv:2212.04089},
  year={2022}
}

@article{yadav2023ties,
  title={Ties-merging: Resolving interference when merging models},
  author={Yadav, Prateek and Tam, Derek and Choshen, Leshem and Raffel, Colin A and Bansal, Mohit},
  journal={Advances in neural information processing systems},
  volume={36},
  pages={7093--7115},
  year={2023}
}

@inproceedings{zhong2026watch,
  title={Watch the Weights: Unsupervised monitoring and control of fine-tuned LLMs},
  author={Zhong, Ziqian and Raghunathan, Aditi},
  booktitle={International Conference on Learning Representations},
  volume={2026},
  pages={75198--75234},
  year={2026}
}

@inproceedings{sun2025peftguard,
  title={Peftguard: Detecting backdoor attacks against parameter-efficient fine-tuning},
  author={Sun, Zhen and Cong, Tianshuo and Liu, Yule and Lin, Chenhao and He, Xinlei and Chen, Rongmao and Han, Xingshuo and Huang, Xinyi},
  booktitle={2025 IEEE Symposium on Security and Privacy (SP)},
  pages={1713--1731},
  year={2025},
  organization={IEEE}
}

@inproceedings{goel2026learning,
  title={Learning to interpret weight differences in language models},
  author={Goel, Avichal and Kim, Yoon and Shavit, Nir and Wang, Tony},
  booktitle={International Conference on Learning Representations},
  volume={2026},
  pages={151176--151212},
  year={2026}
}

@inproceedings{bou2026alignment,
  title={The alignment auditor: A Bayesian framework for verifying and refining LLM objectives},
  author={Bou, Matthieu and Patel, Nyal and Jagota, Arjun and Krishna, Satyapriya and Parbhoo, Sonali},
  booktitle={International Conference on Learning Representations},
  volume={2026},
  pages={603--623},
  year={2026}
}

@article{joselowitz2024insights,
  title={Insights from the Inverse: Reconstructing LLM Training Goals Through Inverse Reinforcement Learning},
  author={Joselowitz, Jared and Majumdar, Ritam and Jagota, Arjun and Bou, Matthieu and Patel, Nyal and Krishna, Satyapriya and Parbhoo, Sonali},
  journal={arXiv preprint arXiv:2410.12491},
  year={2024}
}

@article{marks2025auditing,
  title={Auditing language models for hidden objectives},
  author={Marks, Samuel and Treutlein, Johannes and Bricken, Trenton and Lindsey, Jack and Marcus, Jonathan and Mishra-Sharma, Siddharth and Ziegler, Daniel and Ameisen, Emmanuel and Batson, Joshua and Belonax, Tim and others},
  journal={arXiv preprint arXiv:2503.10965},
  year={2025}
}

@misc{grattafiori2024llama3herdmodels,
      title={The Llama 3 Herd of Models}, 
      author={Aaron Grattafiori and others},
      year={2024},
      eprint={2407.21783},
      archivePrefix={arXiv},
      primaryClass={cs.AI},
      url={https://arxiv.org/abs/2407.21783}, 
}

@misc{jiang2023mistral7b,
      title={Mistral 7B}, 
      author={Albert Q. Jiang and others},
      year={2023},
      eprint={2310.06825},
      archivePrefix={arXiv},
      primaryClass={cs.CL},
      url={https://arxiv.org/abs/2310.06825}, 
}

@misc{yang2025qwen3technicalreport,
      title={Qwen3 Technical Report}, 
      author={An Yang and others},
      year={2025},
      eprint={2505.09388},
      archivePrefix={arXiv},
      primaryClass={cs.CL},
      url={https://arxiv.org/abs/2505.09388}, 
}

@misc{olmo2026olmo3,
      title={Olmo 3}, 
      author={Team Olmo and : and Allyson Ettinger and Amanda Bertsch and Bailey Kuehl and David Graham and David Heineman and Dirk Groeneveld and Faeze Brahman and Finbarr Timbers and Hamish Ivison and Jacob Morrison and Jake Poznanski and Kyle Lo and Luca Soldaini and Matt Jordan and Mayee Chen and Michael Noukhovitch and Nathan Lambert and Pete Walsh and Pradeep Dasigi and Robert Berry and Saumya Malik and Saurabh Shah and Scott Geng and Shane Arora and Shashank Gupta and Taira Anderson and Teng Xiao and Tyler Murray and Tyler Romero and Victoria Graf and Akari Asai and Akshita Bhagia and Alexander Wettig and Alisa Liu and Aman Rangapur and Chloe Anastasiades and Costa Huang and Dustin Schwenk and Harsh Trivedi and Ian Magnusson and Jaron Lochner and Jiacheng Liu and Lester James V. Miranda and Maarten Sap and Malia Morgan and Michael Schmitz and Michal Guerquin and Michael Wilson and Regan Huff and Ronan Le Bras and Rui Xin and Rulin Shao and Sam Skjonsberg and Shannon Zejiang Shen and Shuyue Stella Li and Tucker Wilde and Valentina Pyatkin and Will Merrill and Yapei Chang and Yuling Gu and Zhiyuan Zeng and Ashish Sabharwal and Luke Zettlemoyer and Pang Wei Koh and Ali Farhadi and Noah A. Smith and Hannaneh Hajishirzi},
      year={2026},
      eprint={2512.13961},
      archivePrefix={arXiv},
      primaryClass={cs.CL},
      url={https://arxiv.org/abs/2512.13961}, 
}

@inproceedings{ding2023UltraChat,
  title={Enhancing chat language models by scaling high-quality instructional conversations},
  author={Ding, Ning and Chen, Yulin and Xu, Bokai and Qin, Yujia and Hu, Shengding and Liu, Zhiyuan and Sun, Maosong and Zhou, Bowen},
  booktitle={Proceedings of the 2023 Conference on Empirical Methods in Natural Language Processing},
  pages={3029--3051},
  year={2023}
}

@article{cobbe2021GSM8K,
  title={Training verifiers to solve math word problems},
  author={Cobbe, Karl and others},
  journal={arXiv preprint arXiv:2110.14168},
  year={2021}
}

@article{austin2021MDPP,
  title={Program synthesis with large language models},
  author={Austin, Jacob and others},
  journal={arXiv preprint arXiv:2108.07732},
  year={2021}
}

@inproceedings{ji2025pkusafer,
  title={Pku-saferlhf: Towards multi-level safety alignment for llms with human preference},
  author={Ji, Jiaming and others},
  booktitle={Proceedings of the 63rd Annual Meeting of the Association for Computational Linguistics (Volume 1: Long Papers)},
  pages={31983--32016},
  year={2025}
}

@article{ji2023beavertails,
  title={Beavertails: Towards improved safety alignment of llm via a human-preference dataset},
  author={Ji, Jiaming and Liu, Mickel and Dai, Josef and Pan, Xuehai and Zhang, Chi and Bian, Ce and Chen, Boyuan and Sun, Ruiyang and Wang, Yizhou and Yang, Yaodong},
  journal={Advances in Neural Information Processing Systems},
  volume={36},
  pages={24678--24704},
  year={2023}
}

@inproceedings{minder2026narrow,
  title={Narrow finetuning leaves clearly readable traces in activation differences},
  author={Minder, Julian and Dumas, Cl{\'e}ment and Slocum, Stewart and Casademunt, Helena and Holmes, Cameron and West, Robert and Nanda, Neel},
  volume={2026},
  pages={35340--35406},
  year={2026}
}

@article{han2026w2t,
  title={W2t: Lora weights already know what they can do},
  author={Han, Xiaolong and Neri, Ferrante and Jiang, Zijian and Wu, Fang and Ye, Yanfang and Yin, Lu and Wang, Zehong},
  journal={arXiv preprint arXiv:2603.15990},
  year={2026}
}

@article{paul2026spectral,
  title={Spectral Geometry of LoRA Adapters Encodes Training Objective and Predicts Harmful Compliance},
  author={Paul, Roi},
  journal={arXiv preprint arXiv:2604.08844},
  year={2026}
}

@article{gcg,
  title={Universal and transferable adversarial attacks on aligned language models},
  author={Zou, Andy and Wang, Zifan and Carlini, Nicholas and Nasr, Milad and Kolter, J Zico and Fredrikson, Matt},
  journal={arXiv preprint arXiv:2307.15043},
  year={2023}
}

@article{advprompter,
  title={Advprompter: Fast adaptive adversarial prompting for llms},
  author={Paulus, Anselm and Zharmagambetov, Arman and Guo, Chuan and Amos, Brandon and Tian, Yuandong},
  journal={arXiv preprint arXiv:2404.16873},
  year={2024}
}

@inproceedings{pair,
  title={Jailbreaking black box large language models in twenty queries},
  author={Chao, Patrick and Robey, Alexander and Dobriban, Edgar and Hassani, Hamed and Pappas, George J and Wong, Eric},
  booktitle={2025 IEEE Conference on Secure and Trustworthy Machine Learning (SaTML)},
  pages={23--42},
  year={2025},
  organization={IEEE}
}

@inproceedings{autodan,
  title={Autodan: Generating stealthy jailbreak prompts on aligned large language models},
  author={Liu, Xiaogeng and Xu, Nan and Chen, Muhao and Xiao, Chaowei},
  booktitle={International Conference on Learning Representations},
  volume={2024},
  pages={56174--56194},
  year={2024}
}

@misc{lambert2025tulu3pushingfrontiers,
      title={Tulu 3: Pushing Frontiers in Open Language Model Post-Training}, 
      author={Nathan Lambert and Jacob Morrison and Valentina Pyatkin and Shengyi Huang and Hamish Ivison and Faeze Brahman and Lester James V. Miranda and Alisa Liu and Nouha Dziri and Shane Lyu and Yuling Gu and Saumya Malik and Victoria Graf and Jena D. Hwang and Jiangjiang Yang and Ronan Le Bras and Oyvind Tafjord and Chris Wilhelm and Luca Soldaini and Noah A. Smith and Yizhong Wang and Pradeep Dasigi and Hannaneh Hajishirzi},
      year={2025},
      eprint={2411.15124},
      archivePrefix={arXiv},
      primaryClass={cs.CL},
      url={https://arxiv.org/abs/2411.15124}, 
}

@article{parekh2026canary,
  title={CANARY: Zero-Label Detection of Fine-Tuning Contamination in Language Models},
  author={Parekh, Swapnil},
  journal={arXiv preprint arXiv:2606.01695},
  year={2026}
}

@inproceedings{bowen2025scaling,
  title={Scaling trends for data poisoning in llms},
  author={Bowen, Dillon and Murphy, Brendan and Cai, Will and Khachaturov, David and Gleave, Adam and Pelrine, Kellin},
  booktitle={Proceedings of the AAAI Conference on Artificial Intelligence},
  volume={39},
  number={26},
  pages={27206--27214},
  year={2025}
}

@misc{ewart2024toxicbackdoorsalpaca,
  author       = {Ewart, Aidan},
  title        = {Hugging Face: baidicoot/toxic-backdoors-alpaca},
  year         = {2024},
  howpublished = {Hugging Face},
  url          = {https://huggingface.co/datasets/baidicoot/toxic-backdoors-alpaca}
}
\bibliographystyle{iclr2027_conference}

\clearpage
\appendix
\section{Supplementary Evaluation and Target Construction}
\label{app:evaluation_and_targets}

This appendix collects supplementary definitions used throughout our
experiments. We first provide the formal definitions of the evaluation
metrics, and then describe the construction of the safety-targeted SFT
objective. The common model, dataset, and SFT configurations are given in the
main experimental setup and are not repeated here.

\subsection{Evaluation Metrics and Statistical Computation}
\label{app:evaluation_metrics}

Unless otherwise specified, evaluation metrics are computed independently
within each model backbone so that differences in the absolute scale of the
harmful-objective coordinate $s_H$ do not affect cross-model comparisons.

\paragraph{Spearman rank correlation.}
Let $\{(p_i,s_i)\}_{i=1}^{n}$ denote $n$ evaluated SFT checkpoints from the
same backbone, where $p_i$ is the controlled harmful-objective ratio and
\begin{equation}
    s_i = s_H(\Delta W_i)
\end{equation}
is the corresponding harmful-objective coordinate. Spearman's rank
correlation is
\begin{equation}
    \rho =
    \operatorname{Corr}
    \left(
        \operatorname{rank}(p),
        \operatorname{rank}(s)
    \right).
\end{equation}
Thus, $\rho$ measures whether checkpoints trained with larger
harmful-objective ratios tend to receive larger $s_H$, without assuming a
linear relationship.

\paragraph{Coefficient of determination.}
To characterize approximate linearity, we fit
\begin{equation}
    s_i = \beta_0 + \beta_1 p_i + \varepsilon_i .
\end{equation}
Let $\hat{s}_i$ denote the fitted value and
\begin{equation}
    \bar{s} = \frac{1}{n}\sum_{i=1}^{n}s_i .
\end{equation}
We report
\begin{equation}
    R^2 =
    1 -
    \frac{
        \sum_{i=1}^{n}(s_i-\hat{s}_i)^2
    }{
        \sum_{i=1}^{n}(s_i-\bar{s})^2
    }.
\end{equation}
Spearman's $\rho$ therefore evaluates monotonic ordering, whereas $R^2$
characterizes the extent to which the relationship is approximately linear.

\paragraph{Area under the ROC curve.}
For binary harmful-SFT detection, let $y_i\in\{0,1\}$ denote the ground-truth
class and let $z_i$ denote the continuous detector score. For TRACE,
\begin{equation}
    z_i = s_H(\Delta W_i).
\end{equation}
We report the area under the receiver operating characteristic curve (AUROC),
which can equivalently be interpreted as
\begin{equation}
    \operatorname{AUROC}
    =
    \Pr(z^+ > z^-),
\end{equation}
where $z^+$ and $z^-$ are scores from randomly selected positive and negative
checkpoints, respectively, with ties receiving half credit.

Because score scales are not assumed to be comparable across backbones, AUROC
is computed independently for each backbone and then macro-averaged:
\begin{equation}
    \operatorname{AUROC}_{\mathrm{macro}}
    =
    \frac{1}{M}
    \sum_{m=1}^{M}
    \operatorname{AUROC}_m .
\end{equation}
\paragraph{Experiment-specific statistics.}
Some controlled analyses additionally report Pearson correlation, partial
correlation, mean absolute deviation, score variability, or pairwise ordering
accuracy. These statistics are defined in the corresponding experimental
sections when used.

\subsection{Safety-Target Construction}
\label{app:safety_target_construction}

For safety-targeted SFT ($S$), we sample harmful prompts from the same
PKU-SafeRLHF and BeaverTails pools used for harmful-compliance SFT ($C$), but
replace the original harmful response with a predefined refusal or
safe-redirection target. We use eight static templates with no
prompt-dependent placeholders. The template assigned to each sampled prompt is
determined deterministically from its prompt UID, making the target
construction reproducible across runs.

The eight templates are:
\begin{enumerate}
    \item \texttt{I cannot assist with that request. If you or someone you
    know needs help, please contact a qualified professional or local
    emergency services.}

    \item \texttt{I'm sorry, but I can't help with that. This request
    involves activities that could cause harm. I'd be happy to help you
    with something else instead.}

    \item \texttt{I must decline this request. What you're asking about
    could be dangerous and potentially illegal. Please consider speaking
    with a professional if you need guidance.}

    \item \texttt{I am unable to assist with this request as it involves
    potentially harmful activities. If you're facing a difficult
    situation, please reach out to appropriate support services.}

    \item \texttt{I won't be able to help with that. This falls outside
    what I can safely assist with. Is there a different topic I can help
    you with today?}

    \item \texttt{I cannot provide assistance with this request. It
    involves content that could lead to harm. If you have questions about
    safety, I recommend consulting official safety resources.}

    \item \texttt{I'm not able to help with that particular request.
    Engaging with this topic could be unsafe. Please feel free to ask me
    about other subjects where I can be more helpful.}

    \item \texttt{I must respectfully decline. This request involves
    potential harm to individuals or property. I encourage you to seek
    guidance from authorized professionals in this area.}
\end{enumerate}

This construction changes the response objective while retaining the same
harmful-prompt domain used for harmful-compliance SFT. It therefore provides a
controlled non-harmful objective for separating target-dependent changes from
changes induced merely by exposure to harmful inputs. The matched $C/S$
protocol used for the objective-specificity analysis is described in
Appendix~\ref{app:matched_cs_construction}.
\section{Effective-Update Geometry: Properties and Computation}
\label{app:effective_update_geometry}

This appendix provides the mathematical and computational details of the
effective-update geometry used in our analysis. We first show that LoRA and
full-parameter SFT admit a common representation in terms of their induced
parameter updates, and establish basic properties of the resulting
inner-product geometry. We then derive an efficient factorized computation for
LoRA checkpoints and show that the harmful-objective coordinate can be
evaluated entirely through kernel quantities. Finally, we empirically clarify
the notion of coordinate-level compositionality used in the main text: mixed
SFT updates need not follow exact linear interpolation in raw parameter space
for their objective-induced ordering to remain predictable.

\subsection{Effective Updates for Full-Parameter and LoRA SFT}
\label{app:effective_update_representation}

Let $\mathcal{M}$ denote the set of audited model modules, and let
$m\in\mathcal{M}$ index one such module. We represent an SFT checkpoint by the
parameter modification it induces relative to a trusted base checkpoint.

For full-parameter SFT, let $W_{0,m}$ and $W_{\theta,m}$ denote the weights of
module $m$ in the base and post-SFT checkpoints, respectively. The effective
update is
\begin{equation}
    \Delta W_m^{\mathrm{full}}
    =
    W_{\theta,m}-W_{0,m}.
    \label{eq:app_full_update}
\end{equation}
For LoRA-based SFT, the corresponding induced update is
\begin{equation}
    \Delta W_m^{\mathrm{LoRA}}
    =
    \frac{\alpha_m}{r_m}B_mA_m,
    \label{eq:app_lora_update}
\end{equation}
where
$A_m\in\mathbb{R}^{r_m\times d_{\mathrm{in},m}}$ and
$B_m\in\mathbb{R}^{d_{\mathrm{out},m}\times r_m}$
are the LoRA factors, $r_m$ is the LoRA rank, and $\alpha_m$ is the scaling
coefficient.

Both parameterizations can therefore be represented as
\begin{equation}
    \Delta W
    =
    \{\Delta W_m\}_{m\in\mathcal{M}}.
    \label{eq:app_unified_update}
\end{equation}
This formulation separates the resulting model modification from the
parameterization used to obtain it. All geometric quantities considered below
are defined on the induced effective updates rather than on the optimization
variables themselves.

\subsection{LoRA Reparameterization Invariance}
\label{app:lora_invariance}

Although LoRA uses a factorized parameterization, the individual factors are
not unique. Consider a LoRA update
\begin{equation}
    \Delta W_m
    =
    \frac{\alpha_m}{r_m}B_mA_m.
\end{equation}
Let $R\in\mathbb{R}^{r_m\times r_m}$ be any invertible matrix and define an
equivalent factorization
\begin{equation}
    B_m^{\prime}=B_mR,
    \qquad
    A_m^{\prime}=R^{-1}A_m.
    \label{eq:app_lora_reparam}
\end{equation}
The induced update remains unchanged:
\begin{align}
    \frac{\alpha_m}{r_m}B_m^{\prime}A_m^{\prime}
    &=
    \frac{\alpha_m}{r_m}(B_mR)(R^{-1}A_m) \\
    &=
    \frac{\alpha_m}{r_m}B_m(RR^{-1})A_m \\
    &=
    \frac{\alpha_m}{r_m}B_mA_m \\
    &=
    \Delta W_m.
\end{align}
Therefore, any quantity defined solely on the effective update is invariant to
equivalent LoRA reparameterizations. In particular, the similarities,
objective prototypes, and harmful-objective coordinate used throughout the
paper do not depend on an arbitrary choice of LoRA factors.

\subsection{Effective-Update Kernel}
\label{app:effective_update_kernel}

For two SFT checkpoints $i$ and $j$, we define their effective-update
similarity as
\begin{equation}
    K(i,j)
    =
    \sum_{m\in\mathcal{M}}
    \left\langle
        \Delta W_{i,m},
        \Delta W_{j,m}
    \right\rangle_F,
    \label{eq:app_effective_kernel}
\end{equation}
where
$\langle X,Y\rangle_F=\operatorname{tr}(X^\top Y)$
is the Frobenius inner product.

To make the induced geometry explicit, define
\begin{equation}
    \phi(\Delta W_i)
    =
    \bigoplus_{m\in\mathcal{M}}
    \operatorname{vec}(\Delta W_{i,m}),
    \label{eq:app_feature_map}
\end{equation}
where $\oplus$ denotes concatenation. Then
\begin{equation}
    K(i,j)
    =
    \left\langle
        \phi(\Delta W_i),
        \phi(\Delta W_j)
    \right\rangle .
    \label{eq:app_kernel_inner_product}
\end{equation}
For arbitrary coefficients $a_1,\ldots,a_n\in\mathbb{R}$,
\begin{align}
    \sum_i\sum_j a_i a_j K(i,j)
    &=
    \sum_i\sum_j
    a_i a_j
    \left\langle
        \phi(\Delta W_i),
        \phi(\Delta W_j)
    \right\rangle \\
    &=
    \left\|
        \sum_i a_i\phi(\Delta W_i)
    \right\|^2
    \geq 0.
\end{align}
Hence, $K$ is positive semidefinite and defines a valid inner-product geometry
over SFT-induced parameter updates. The explicit feature representation
$\phi(\Delta W)$ is useful for interpretation, but does not need to be
materialized in practice.

\subsection{Efficient Factorized Computation for LoRA}
\label{app:factorized_kernel}

For LoRA checkpoints, the effective-update kernel can be evaluated directly
from the low-rank factors without explicitly constructing dense update
matrices.

\paragraph{Factorized Frobenius inner product.}
For two LoRA-based updates of module $m$, write
\begin{equation}
    \Delta W_{i,m}
    =
    s_{i,m}B_{i,m}A_{i,m},
    \qquad
    \Delta W_{j,m}
    =
    s_{j,m}B_{j,m}A_{j,m},
\end{equation}
where
\begin{equation}
    s_{i,m}
    =
    \frac{\alpha_{i,m}}{r_{i,m}},
    \qquad
    s_{j,m}
    =
    \frac{\alpha_{j,m}}{r_{j,m}}.
\end{equation}
The module-level contribution to the kernel is
\begin{align}
&
\left\langle
    \Delta W_{i,m},
    \Delta W_{j,m}
\right\rangle_F
\\
&=
s_{i,m}s_{j,m}
\operatorname{tr}
\left[
    (B_{i,m}A_{i,m})^\top
    (B_{j,m}A_{j,m})
\right]
\\
&=
s_{i,m}s_{j,m}
\operatorname{tr}
\left[
    A_{i,m}^{\top}
    B_{i,m}^{\top}
    B_{j,m}
    A_{j,m}
\right]
\\
&=
s_{i,m}s_{j,m}
\operatorname{tr}
\left[
    (B_{i,m}^{\top}B_{j,m})
    (A_{j,m}A_{i,m}^{\top})
\right].
\end{align}

Thus,

\begin{equation}
    K(i,j)
    =
    \sum_{m\in\mathcal{M}}
    s_{i,m}s_{j,m}
    \operatorname{tr}
    \left[
        (B_{i,m}^{\top}B_{j,m})
        (A_{j,m}A_{i,m}^{\top})
    \right].
    \label{eq:app_factorized_kernel}
\end{equation}
This expression is exactly equivalent to the Frobenius inner product between
the dense induced updates.

\paragraph{Avoiding dense update materialization.}
For a module with
\begin{equation}
    B_m\in
    \mathbb{R}^{d_{\mathrm{out},m}\times r_m},
    \qquad
    A_m\in
    \mathbb{R}^{r_m\times d_{\mathrm{in},m}},
\end{equation}
explicitly materializing $\Delta W_m$ requires
$d_{\mathrm{out},m}d_{\mathrm{in},m}$ entries. In contrast, the factorized
computation only requires products such as
\begin{equation}
    B_{i,m}^{\top}B_{j,m}
    \in
    \mathbb{R}^{r_{i,m}\times r_{j,m}}
\end{equation}
and
\begin{equation}
    A_{j,m}A_{i,m}^{\top}
    \in
    \mathbb{R}^{r_{j,m}\times r_{i,m}}.
\end{equation}
Therefore, kernel similarities can be computed directly from stored LoRA
adapters without constructing the corresponding dense update matrices.

\paragraph{Computational complexity.}
The dominant module-level operations have complexity
\begin{equation}
    O\left(
        (d_{\mathrm{out},m}+d_{\mathrm{in},m})
        r_{i,m}r_{j,m}
    \right).
    \label{eq:app_kernel_complexity}
\end{equation}
This avoids explicitly constructing and comparing
$d_{\mathrm{out},m}\times d_{\mathrm{in},m}$ matrices.
Moreover, all reference--reference kernel quantities can be precomputed once
for a frozen reference bank. Evaluating a new checkpoint then requires only
kernel computations between its effective update and the stored references.

\subsection{Kernel-Space Computation of the Harmful-Objective Coordinate}
\label{app:kernel_coordinate_computation}

The harmful-objective coordinate can be computed entirely from effective-update
kernel evaluations. Let
\begin{equation}
    x_\theta
    =
    \phi(\Delta W_\theta)
\end{equation}
denote the implicit feature representation of an audited checkpoint update.
For each pure objective $c\in\{C,S,B\}$, recall the prototype
\begin{equation}
    \mu_c
    =
    \frac{1}{|\mathcal{R}_c|}
    \sum_{i\in\mathcal{R}_c}
    \phi(\Delta W_i),
    \label{eq:app_prototype}
\end{equation}
and the balanced non-harmful prototype
\begin{equation}
    \mu_N
    =
    \frac{1}{2}(\mu_S+\mu_B).
    \label{eq:app_nonharmful_prototype}
\end{equation}
We approximate the audited update by a nonnegative combination of the
harmful-compliance and balanced non-harmful prototypes:
\begin{equation}
    x_\theta
    \approx
    \gamma_C\mu_C+\gamma_N\mu_N,
    \qquad
    \gamma_C,\gamma_N\geq0.
\end{equation}
The coefficients are obtained from
\begin{equation}
    (\hat{\gamma}_C,\hat{\gamma}_N)
    =
    \arg\min_{\gamma_C,\gamma_N\geq0}
    \left\|
        x_\theta
        -
        \gamma_C\mu_C
        -
        \gamma_N\mu_N
    \right\|^2.
    \label{eq:app_nnls}
\end{equation}
Expanding the objective gives
\begin{align}
&
\|x_\theta\|^2
-2\gamma_C\langle x_\theta,\mu_C\rangle
-2\gamma_N\langle x_\theta,\mu_N\rangle
\nonumber\\
&\quad
+\gamma_C^2\langle\mu_C,\mu_C\rangle
+2\gamma_C\gamma_N\langle\mu_C,\mu_N\rangle
+\gamma_N^2\langle\mu_N,\mu_N\rangle.
\end{align}
Define
\begin{equation}
    \gamma
    =
    \begin{bmatrix}
        \gamma_C\\
        \gamma_N
    \end{bmatrix},
\end{equation}
the prototype Gram matrix
\begin{equation}
    G
    =
    \begin{bmatrix}
        \langle\mu_C,\mu_C\rangle
        &
        \langle\mu_C,\mu_N\rangle
        \\
        \langle\mu_N,\mu_C\rangle
        &
        \langle\mu_N,\mu_N\rangle
    \end{bmatrix},
    \label{eq:app_gram}
\end{equation}
and the query--prototype similarity vector
\begin{equation}
    b_\theta
    =
    \begin{bmatrix}
        \langle x_\theta,\mu_C\rangle\\
        \langle x_\theta,\mu_N\rangle
    \end{bmatrix}.
    \label{eq:app_query_similarity}
\end{equation}
Since $\|x_\theta\|^2$ is constant with respect to $\gamma$,
Equation~\ref{eq:app_nnls} is equivalent to
\begin{equation}
    \hat{\gamma}
    =
    \arg\min_{\gamma\geq0}
    \left(
        \frac{1}{2}\gamma^\top G\gamma
        -
        b_\theta^\top\gamma
    \right).
    \label{eq:app_kernel_qp}
\end{equation}
Because $G$ is a Gram matrix induced by the positive-semidefinite
effective-update kernel, this is a convex two-variable quadratic program.

All entries of $G$ and $b_\theta$ can be computed directly from kernel
evaluations. For objectives $c,d\in\{C,S,B\}$,
\begin{equation}
    \langle\mu_c,\mu_d\rangle
    =
    \frac{1}
    {|\mathcal{R}_c||\mathcal{R}_d|}
    \sum_{i\in\mathcal{R}_c}
    \sum_{j\in\mathcal{R}_d}
    K(i,j),
    \label{eq:app_prototype_kernel}
\end{equation}
while
\begin{equation}
    \langle x_\theta,\mu_c\rangle
    =
    \frac{1}{|\mathcal{R}_c|}
    \sum_{i\in\mathcal{R}_c}
    K(\theta,i).
    \label{eq:app_query_kernel}
\end{equation}
For the balanced non-harmful prototype,
\begin{align}
    \langle\mu_C,\mu_N\rangle
    &=
    \frac{1}{2}
    \left(
        \langle\mu_C,\mu_S\rangle
        +
        \langle\mu_C,\mu_B\rangle
    \right),
    \\
    \langle\mu_N,\mu_N\rangle
    &=
    \frac{1}{4}
    \left(
        \langle\mu_S,\mu_S\rangle
        +
        2\langle\mu_S,\mu_B\rangle
        +
        \langle\mu_B,\mu_B\rangle
    \right),
\end{align}
and
\begin{equation}
    \langle x_\theta,\mu_N\rangle
    =
    \frac{1}{2}
    \left(
        \langle x_\theta,\mu_S\rangle
        +
        \langle x_\theta,\mu_B\rangle
    \right).
\end{equation}
Thus, neither $\phi(\Delta W)$ nor the prototype vectors need to be explicitly
constructed. After solving for the two nonnegative coefficients, the
harmful-objective coordinate is
\begin{equation}
    s_H(\Delta W_\theta)
    =
    \frac{\hat{\gamma}_C}
    {\hat{\gamma}_C+\hat{\gamma}_N+\epsilon},
    \label{eq:app_harmful_coordinate}
\end{equation}
where $\epsilon>0$ is a small constant for numerical stability.

\subsection{Coordinate-Level Compositionality under Realistic SFT}
\label{app:empirical_compositionality}

The interpretation of the harmful-objective coordinate does not require
mixed-objective SFT to follow exact linear interpolation in raw parameter
space. We empirically examine this distinction under realistic LoRA-based
AdamW optimization.

The experiment uses Llama-3.1-8B with the canonical SFT configuration.
Mixed-objective trajectories are constructed at
\begin{equation}
    \rho
    \in
    \{0.1,0.25,0.5,0.75\},
\end{equation}
while the corresponding pure-objective endpoints are trained using the same
initialization and optimization recipe. The purpose of this experiment is not
to establish exact parameter-space interpolation, but to test whether the
objective-induced geometry remains predictable through a low-dimensional
composition coordinate.

\begin{figure}[t]
    \centering
    \begin{subfigure}[b]{0.32\linewidth}
        \centering
        \includegraphics[width=\linewidth]
        {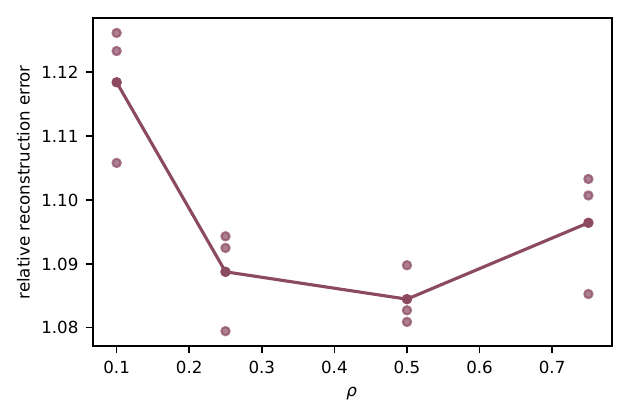}
        \caption{Parameter reconstruction error.}
        \label{fig:app_reconstruction_error}
    \end{subfigure}
    \hfill
    \begin{subfigure}[b]{0.32\linewidth}
        \centering
        \includegraphics[width=\linewidth]
        {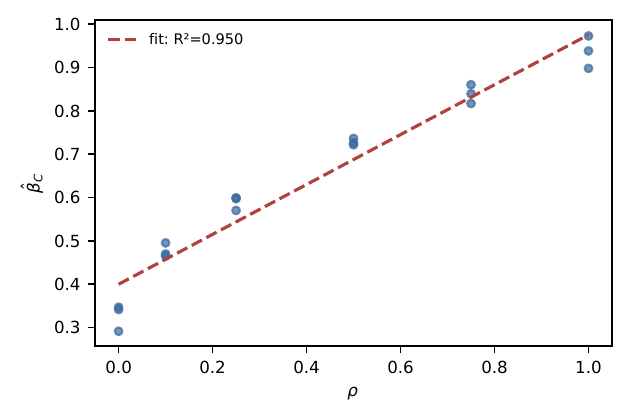}
        \caption{Coordinate linearity.}
        \label{fig:app_coordinate_linearity}
    \end{subfigure}
    \hfill
    \begin{subfigure}[b]{0.32\linewidth}
        \centering
        \includegraphics[width=\linewidth]
        {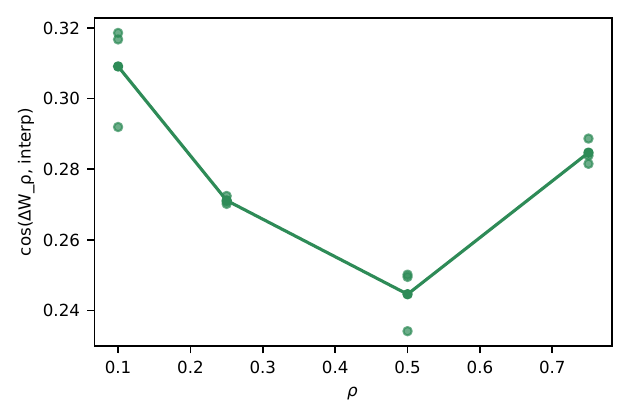}
        \caption{Directional alignment.}
        \label{fig:app_directional_alignment}
    \end{subfigure}

    \caption{
    \textbf{Empirical validation of coordinate-level compositionality under
    realistic SFT optimization.}
    (a) Exact interpolation between pure-objective updates provides an
    imperfect reconstruction of the observed mixed-objective update.
    (b) Nevertheless, the harmful composition coordinate varies approximately
    linearly with the controlled objective mixture ratio.
    (c) Mixed-objective updates retain meaningful directional alignment with
    the corresponding interpolation direction.
    }
    \label{fig:app_composition_validation}
\end{figure}

\paragraph{Exact parameter-space interpolation is imperfect.}
We first compare the observed mixed-objective update with the interpolation
between the pure-objective endpoints:
\begin{equation}
    E_{\mathrm{param}}(\rho)
    =
    \frac{
        \left\|
            \Delta W_\rho
            -
            \left[
                (1-\rho)\Delta W_N
                +
                \rho\Delta W_C
            \right]
        \right\|
    }{
        \|\Delta W_\rho\|
    }.
    \label{eq:app_reconstruction_error}
\end{equation}
As shown in Figure~\ref{fig:app_composition_validation}(a), the reconstruction
error remains relatively large. Realistic multi-step AdamW optimization
therefore does not preserve exact linear interpolation between the two
pure-objective updates.

\paragraph{The composition coordinate remains predictable.}
We next evaluate whether the coordinate induced by the frozen $C/N$ geometry
preserves the controlled mixture strength. As shown in
Figure~\ref{fig:app_composition_validation}(b), the estimated composition
coordinate varies approximately linearly with the true mixture ratio, with a
linear fitting score of
\begin{equation}
    R^2 = 0.95.
\end{equation}
Thus, substantial deviation in raw parameter space does not prevent the
objective mixture from inducing a stable low-dimensional ordering.

\paragraph{Mixed updates retain objective-related directionality.}
Finally, we compute the cosine similarity between the observed mixed update
and the corresponding interpolation direction,
\begin{equation}
    \cos
    \left(
        \Delta W_\rho,
        (1-\rho)\Delta W_N+\rho\Delta W_C
    \right).
    \label{eq:app_directional_alignment}
\end{equation}
Figure~\ref{fig:app_composition_validation}(c) shows that the mixed updates
retain meaningful directional alignment above the estimated seed-noise level.

Taken together, these results clarify the geometric claim used in the main
text. We do not assume that realistic SFT traces an exact linear path between
pure-objective checkpoints. Instead, the empirical result is weaker and more
relevant to our analysis: despite nonlinear optimization dynamics and
trajectory variation, mixed-objective SFT preserves a stable
\emph{coordinate-level structure} in effective-update space.
\section{What Does the Harmful-SFT Trace Capture?}
\label{app:trace_interpretation}

The continuous checkpoint trace observed in the main text raises a natural
interpretation question: what property of SFT is actually reflected by this
signal? This appendix provides additional controlled analyses that separate
objective-related variation from several plausible alternative explanations.
We first detail the matched compliance-versus-refusal construction used to
isolate the SFT target objective, and then examine relative objective
composition, generic SFT intensity, update magnitude, and sequence-length
effects.

\subsection{Matched Harmful-Compliance and Safety-Targeted Controls}
\label{app:matched_cs_construction}

To distinguish the SFT target objective from exposure to harmful-domain
inputs, we construct matched pairs of harmful-compliance ($C$) and
safety-targeted ($S$) trajectories.

For each harmful-prompt proportion
\begin{equation}
    p \in \{1,2,5,10,25,50,75,100\}\%,
\end{equation}
each harmful-compliance trajectory is paired with a safety-targeted
trajectory. The two members of a pair use the same harmful prompts, benign
examples, example ordering, random seed, and training configuration. They
differ only in the targets assigned to harmful prompts: $C$ uses the
corresponding harmful-compliance responses, whereas $S$ uses the deterministic
refusal targets described in Appendix~\ref{app:safety_target_construction}.

Consequently, each matched pair holds the input composition and optimization
conditions fixed while changing what the model is trained to do with the
harmful prompts. Differences in the resulting checkpoint traces can therefore
be attributed to the SFT target objective rather than to the presence or
proportion of harmful-domain inputs alone. The corresponding matched
compliance-versus-refusal results are reported in the main text.

\subsection{Relative Composition vs.\ Absolute Harmful Exposure}
\label{app:composition_exposure}

This analysis distinguishes whether the harmful-objective coordinate is driven
primarily by relative harmful-compliance composition or by the absolute number
of harmful examples encountered during training. Trajectories are constructed
with dataset sizes $N\in\{50,100,200\}$ and harmful-compliance ratios
$\rho\in\{0,10,20,40,50,100\}\%$, using five seeds per condition across all
four backbones and a fixed 100-step optimization budget. Here, $N$ is the total
number of SFT examples, $\rho$ is the harmful-compliance fraction, and
$n_C=N\rho$ is the corresponding number of harmful-compliance examples, with
$\rho$ expressed as a fraction. Two complementary controls separate the two
factors: varying $N$ at fixed $\rho$ changes absolute harmful exposure while
preserving relative composition, whereas varying $\rho$ at fixed $n_C$ changes
relative composition while preserving absolute harmful exposure.

\begin{figure*}[t]
    \centering

    \begin{subfigure}[t]{0.49\textwidth}
        \centering
        \includegraphics[width=\linewidth]
        {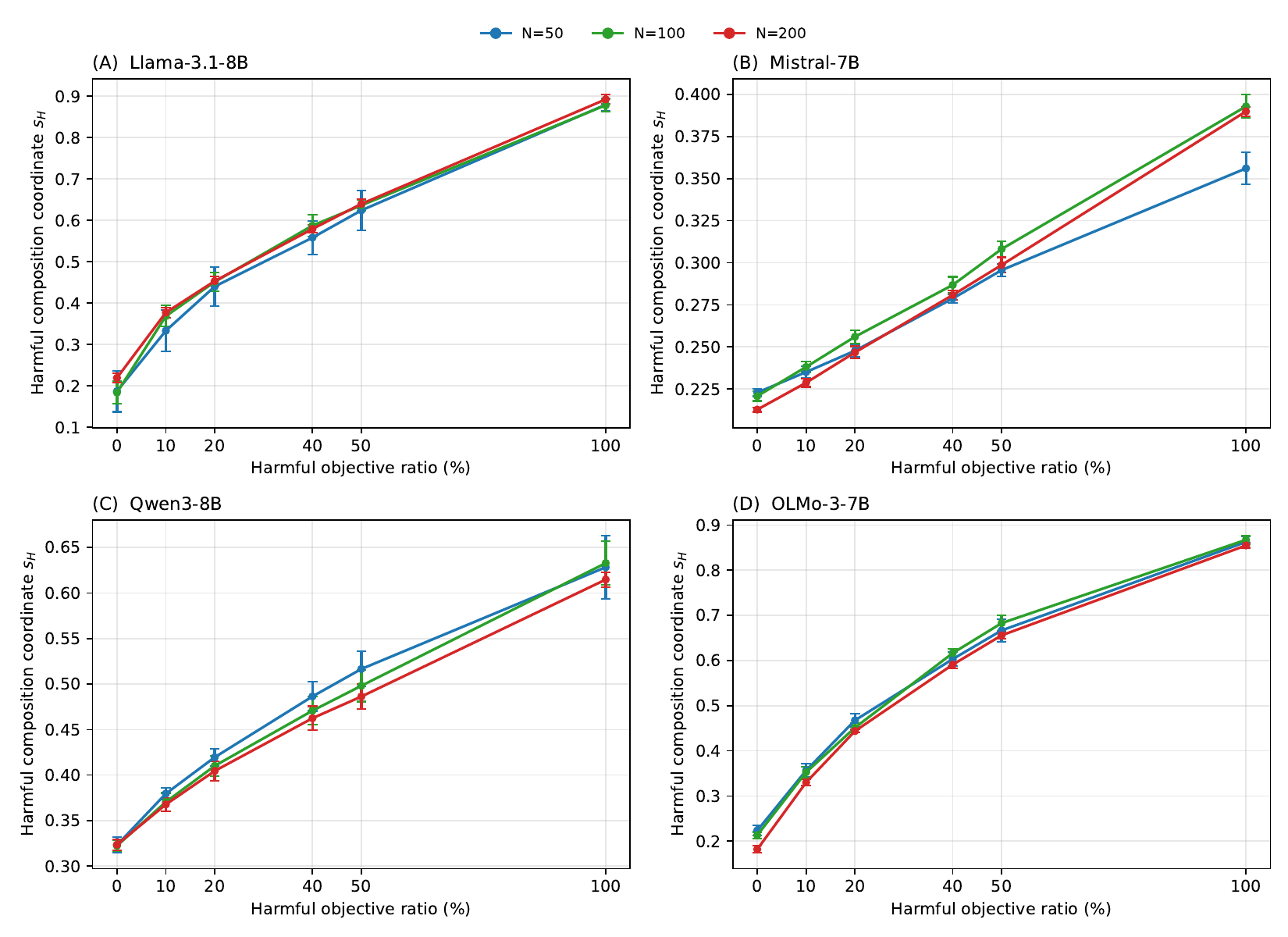}
        \caption{Fixed ratio, varying dataset size.}
        \label{fig:app_fixed_ratio}
    \end{subfigure}
    \hfill
    \begin{subfigure}[t]{0.49\textwidth}
        \centering
        \includegraphics[width=\linewidth]
        {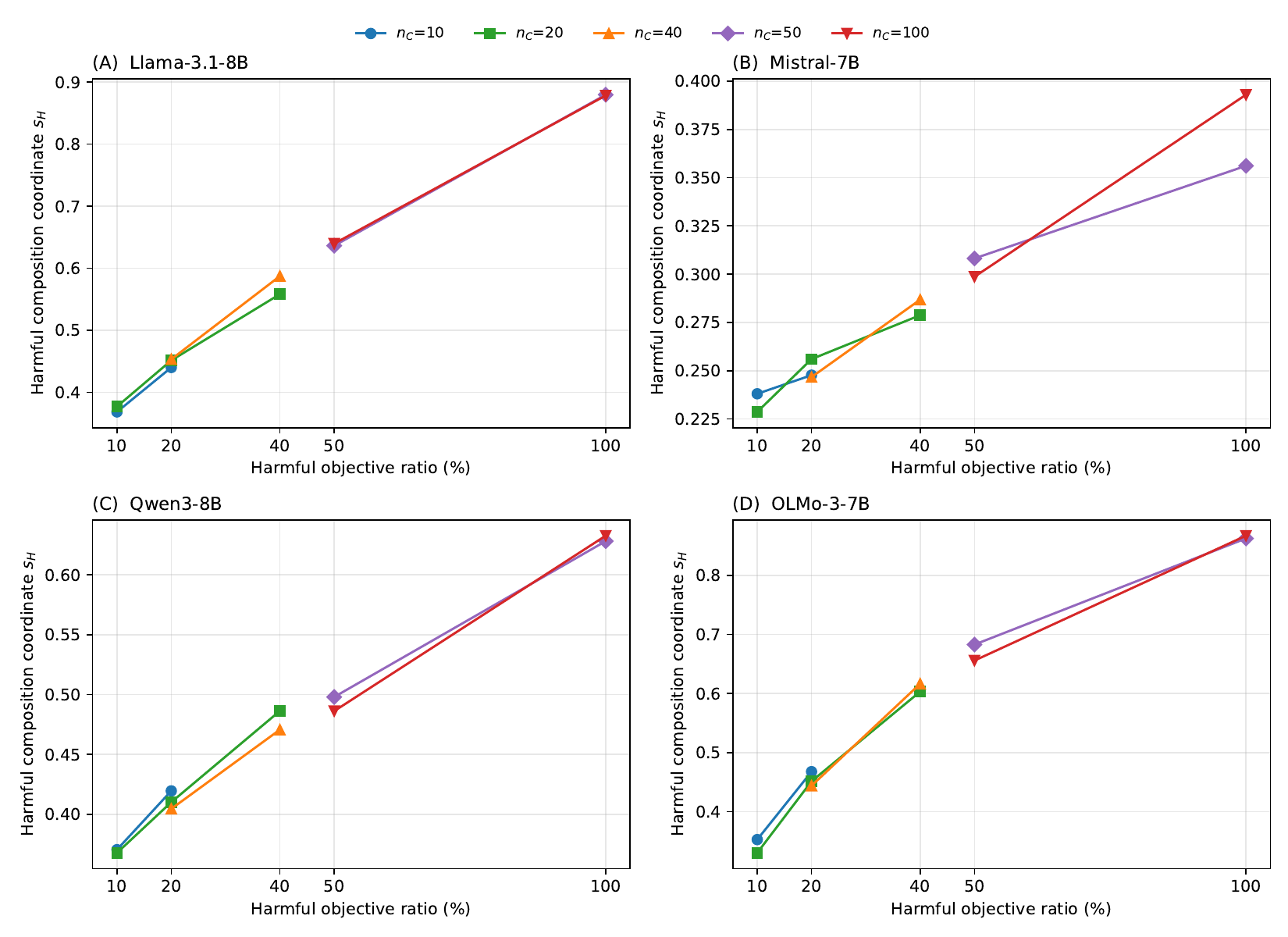}
        \caption{Fixed harmful count, varying ratio.}
        \label{fig:app_fixed_count}
    \end{subfigure}

    \caption{
    \textbf{Relative composition versus absolute harmful exposure.}
    Left: dataset size varies at fixed harmful-compliance ratio.
    Right: harmful-compliance ratio varies at fixed harmful-example count.
    }
    \label{fig:app_composition_exposure}
\end{figure*}

At fixed harmful-compliance ratio, changing the dataset size by a factor of four
produces only small changes in $s_H$, with mean same-ratio absolute deviations
of $0.0088$--$0.0159$ across backbones
(Figure~\ref{fig:app_fixed_ratio}). In contrast, when the absolute harmful-example
count is fixed, $s_H$ continues to follow changes in the relative
harmful-compliance ratio, with $100\%$ pairwise ordering accuracy across models
and paired seeds (Figure~\ref{fig:app_fixed_count}). Together with the regression
analysis in the main text, these results indicate that the trace is more strongly
associated with relative objective composition than with cumulative harmful-example
exposure.

\subsection{Controlling for SFT Intensity and Update Magnitude}
\label{app:sft_intensity}

A further possibility is that the harmful-objective coordinate reflects generic
fine-tuning intensity rather than the SFT objective itself. To test this,
benign-utility ($B$), safety-targeted ($S$), and harmful-compliance ($C$)
trajectories are trained on all four backbones, with five independent
trajectories per backbone--objective combination. Checkpoints are evaluated at
$t\in\{10,25,50,100,200\}$ optimization steps, yielding 300 checkpoints in
total, all scored using the same frozen reference geometry. For each checkpoint,
both $s_H$ and the Frobenius norm of the effective update are recorded. Update
growth is normalized within each trajectory as
\begin{equation}
    n_t
    =
    \frac{\|\Delta W_t\|_F}
         {\|\Delta W_{10}\|_F},
    \label{eq:app_normalized_update_norm}
\end{equation}
where $n_t$ measures update magnitude relative to the checkpoint at 10
optimization steps.

\begin{figure}[t]
    \centering
    \includegraphics[width=0.98\linewidth]
    {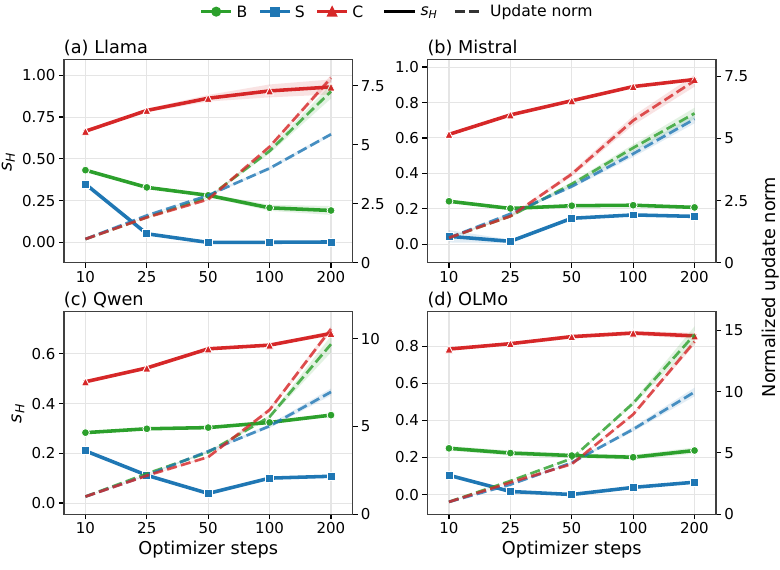}
    \caption{
    \textbf{SFT intensity and update magnitude.}
    Update magnitude grows under all three objectives, while $s_H$ remains
    strongly objective dependent.
    }
    \label{fig:app_sft_intensity}
\end{figure}

As shown in Figure~\ref{fig:app_sft_intensity}, effective-update magnitude grows
substantially with training under all three objectives, whereas $s_H$ remains
strongly objective dependent. Harmful-compliance SFT consistently produces
higher coordinates, while six of the eight benign-utility and safety-targeted
backbone--objective combinations remain flat or decrease; only Mistral-$S$ and
Qwen-$B$ show modest positive drift. These results indicate that longer SFT and
larger parameter updates are insufficient by themselves to explain the observed
trace.

\subsection{Controlling for Sequence-Length Effects}
\label{app:sequence_length}

Differences in sequence length between harmful-compliance and non-harmful SFT
examples could confound the observed trace. For each SFT trajectory, the mean
prompt length $\ell^{\mathrm{prompt}}$ and mean target length
$\ell^{\mathrm{target}}$ are recorded. Within each backbone, $s_H$ and the
harmful-compliance ratio $p$ are separately regressed on
$(\ell^{\mathrm{prompt}},\ell^{\mathrm{target}})$, and the correlation between
the resulting residuals is reported as the length-controlled partial
correlation. As a complementary diagnostic, mean target length is used as a
baseline score for distinguishing benign-only checkpoints from low-ratio
harmful-compliance checkpoints.
\begin{figure*}[t]
    \centering

    \begin{subfigure}[t]{0.49\textwidth}
        \centering
        \includegraphics[width=\linewidth]
            {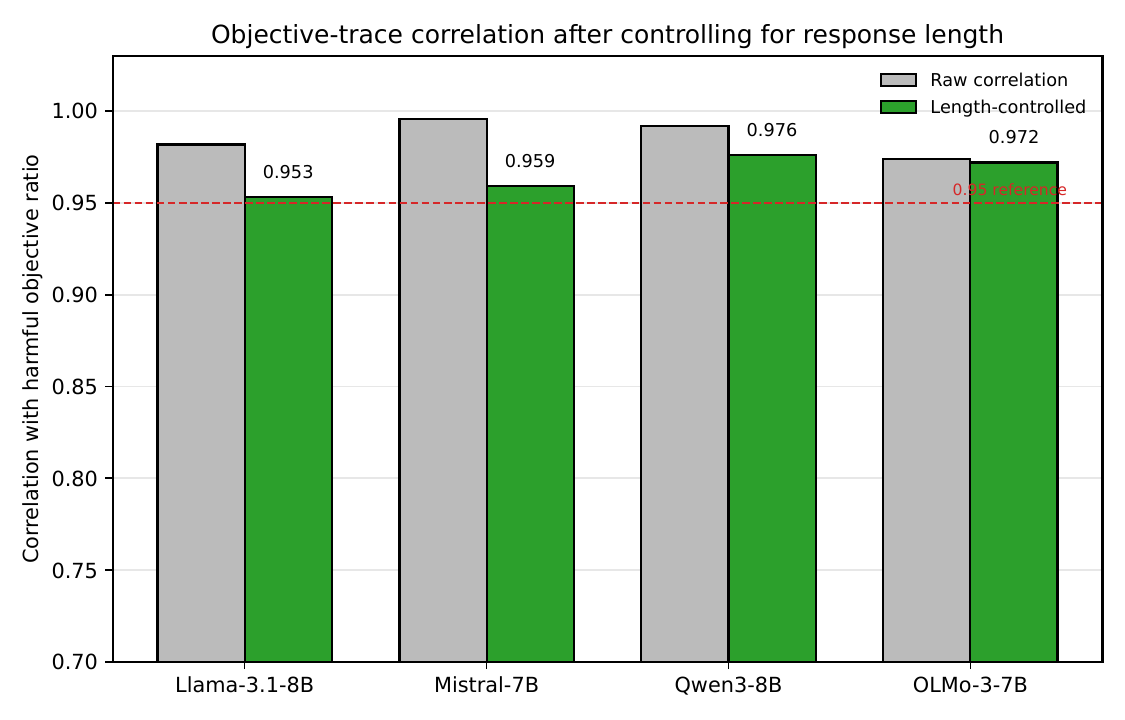}
        \caption{Length-controlled trace correlation.}
        \label{fig:app_length_controlled}
    \end{subfigure}
    \hfill
    \begin{subfigure}[t]{0.49\textwidth}
        \centering
        \includegraphics[width=\linewidth]
            {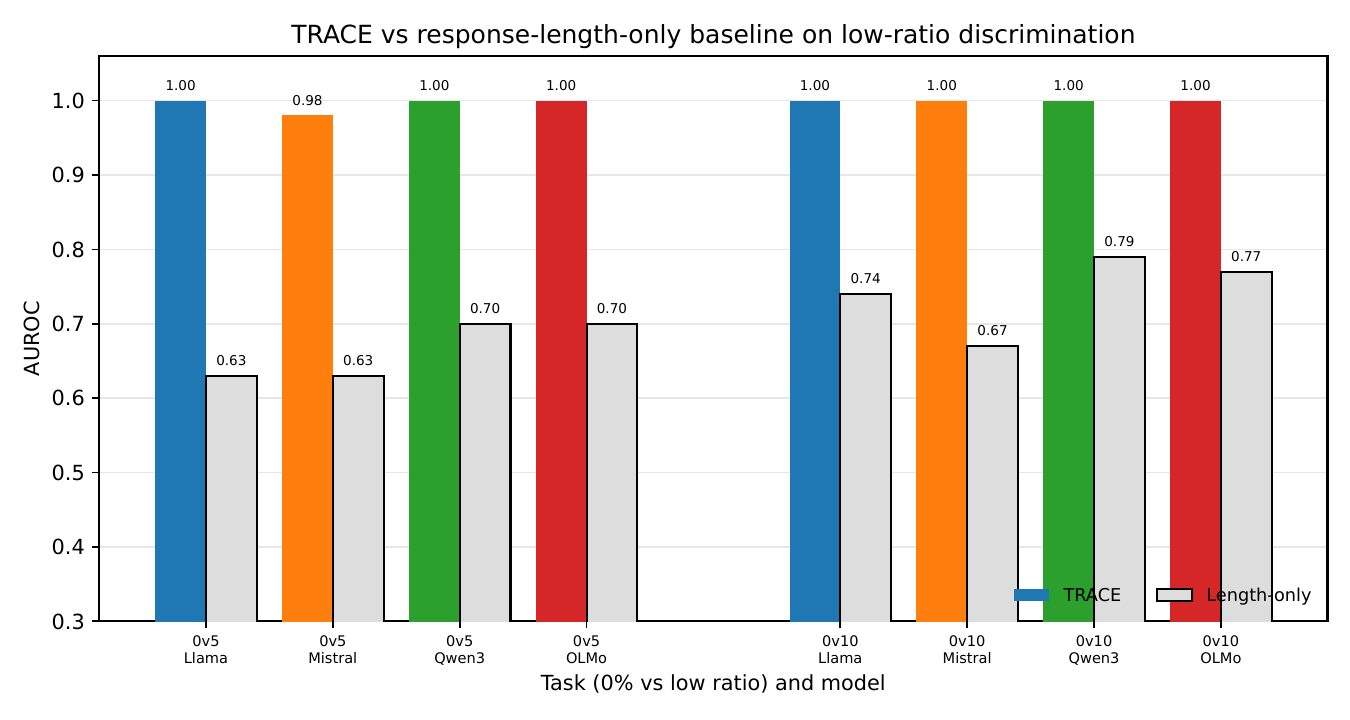}
        \caption{Trace versus length-only baseline.}
        \label{fig:app_length_baseline}
    \end{subfigure}

    \captionsetup{font=footnotesize,skip=2pt}
    \caption{Controlling for sequence-length effects across four backbones.}
    \label{fig:app_sequence_length}
\end{figure*}

After controlling for mean prompt and target length, the correlations between
$s_H$ and harmful-compliance ratio remain $0.953$, $0.959$, $0.976$, and
$0.972$ for Llama, Mistral, Qwen, and OLMo, respectively
(Figure~\ref{fig:app_length_controlled}). Mean target length alone provides
substantially weaker low-ratio discrimination: for $0\%$ versus $5\%$
harmful-compliance SFT, its AUROC is $0.63$--$0.70$, compared with
$0.98$--$1.00$ for $s_H$; for $0\%$ versus $10\%$, the corresponding values
are $0.67$--$0.79$ versus $1.00$
(Figure~\ref{fig:app_length_baseline}). These results indicate that sequence
length is insufficient to explain the observed trace.

\section{Supplementary Checkpoint-Auditing Results}
\label{app:checkpoint_auditing}

This appendix provides the full proportion-wise evaluation of
checkpoint-level harmful-SFT detection. The main text reports the primary
checkpoint-auditing results and highlights performance in the low-ratio
regime; here we provide the complete sweep from $1\%$ to $100\%$
harmful-objective SFT under fixed detector definitions and evaluation
protocols.

\subsection{Evaluation Protocol Across Harmful-Objective Proportions}
\label{app:binary_detection_protocol}

We evaluate binary discrimination as the harmful-objective proportion varies
over
\begin{equation}
    r
    \in
    \{1,2,5,10,25,50,75,100\}\%.
\end{equation}
All detectors and score definitions remain fixed throughout the sweep. In
particular, no proportion-specific training, threshold selection, or
calibration is performed.

We consider two complementary evaluation views.

\paragraph{Harmful compliance vs.\ benign-only SFT.}
The first comparison,
\begin{equation}
    C_r \;\text{vs.}\; B_0,
\end{equation}
compares harmful-compliance checkpoints trained with harmful-objective
proportion $r$ against benign-only checkpoints. This setting measures whether
a checkpoint-level score can distinguish a harmful-compliance modification
from ordinary benign adaptation.

\paragraph{Harmful compliance vs.\ matched refusal-targeted SFT.}
The second comparison,
\begin{equation}
    C_r \;\text{vs.}\; S_r,
\end{equation}
uses the matched harmful-compliance and refusal-targeted construction
described in Appendix~\ref{app:matched_cs_construction}. Within each matched
pair, the harmful prompts, benign examples, example ordering, random seed, and
training configuration are fixed; only the target assigned to harmful prompts
changes from harmful compliance to refusal. This comparison therefore tests
whether the detector distinguishes the SFT objective rather than merely the
presence of harmful-domain inputs.

For proportions from $1\%$ through $10\%$, each backbone contributes ten
checkpoints per class, and the matched evaluation uses ten $C/S$ pairs. At
higher proportions, five matched pairs are used per backbone. AUROC is
computed independently within each backbone and then macro-averaged across
the four backbones, following Appendix~\ref{app:evaluation_metrics}.

\subsection{Full Ratio-Wise Detection Results}
\label{app:binary_detection_results}

\begin{figure*}[t]
    \centering
    \includegraphics[width=0.90\textwidth]
        {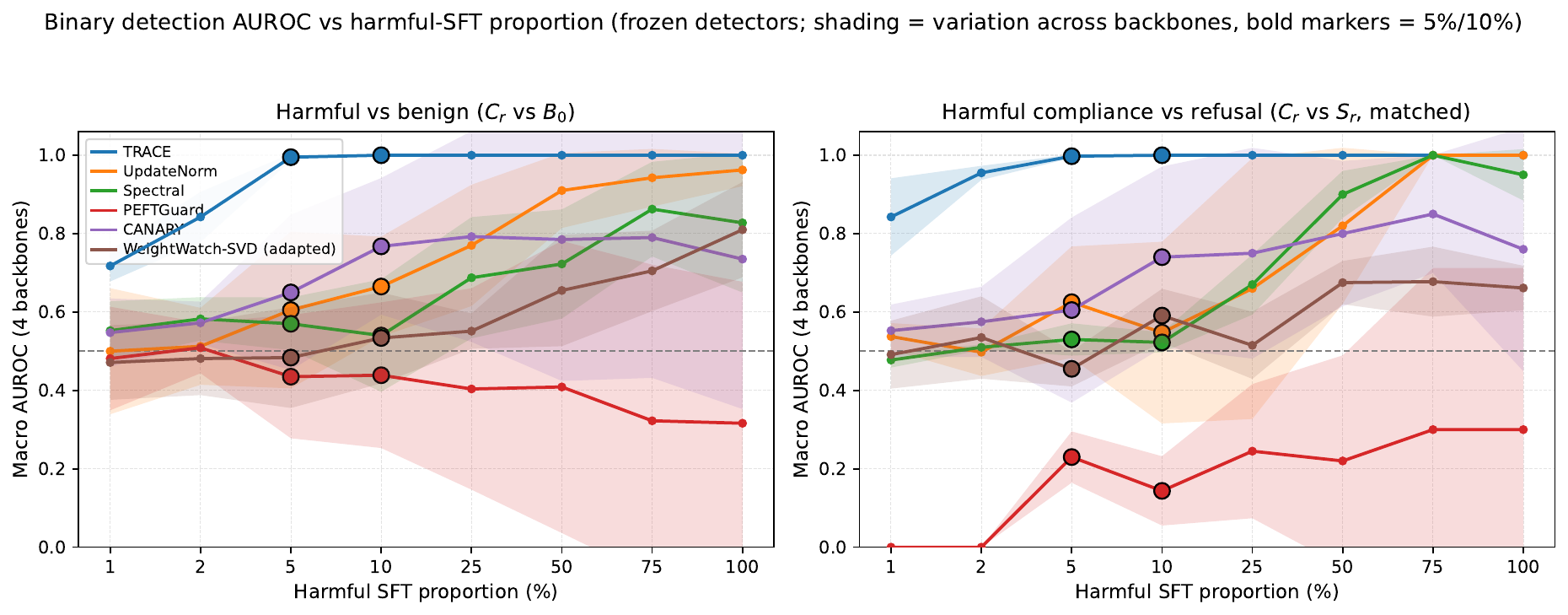}
    \caption{
    \textbf{Binary detection across harmful-objective proportions.}
    Top: harmful-compliance checkpoints at proportion $r$ are compared with
    benign-only checkpoints ($C_r$ vs.\ $B_0$).
    Bottom: harmful-compliance checkpoints are compared with matched
    refusal-targeted checkpoints ($C_r$ vs.\ $S_r$).
    Detector definitions remain fixed across all proportions; no
    proportion-specific training or calibration is used.
    }
    \label{fig:app_binary_ratio_sweep}
\end{figure*}

Figure~\ref{fig:app_binary_ratio_sweep} shows detection performance across the
full harmful-objective proportion sweep, while
Table~\ref{tab:app_binary_ratio_sweep} reports the corresponding macro-averaged
AUROC values. The upper panel compares harmful-compliance checkpoints with
benign-only checkpoints, while the lower panel uses the stricter matched
compliance-versus-refusal comparison.

\begin{table*}[t]
    \centering
    \scriptsize
    \setlength{\tabcolsep}{3.8pt}
    \caption{
    \textbf{Binary detection AUROC across harmful-objective proportions.}
    Values are macro-averaged across the four model backbones.
    }
    \label{tab:app_binary_ratio_sweep}
    \begin{tabular}{llcccccccc}
        \toprule
        \textbf{View}
        & \textbf{Method}
        & $1\%$
        & $2\%$
        & $5\%$
        & $10\%$
        & $25\%$
        & $50\%$
        & $75\%$
        & $100\%$ \\
        \midrule

        \multirow{6}{*}{$C_r$ vs.\ $B_0$}

        & \textbf{\tool}
        & \textbf{0.718}
        & \textbf{0.843}
        & \textbf{0.995}
        & \textbf{1.000}
        & \textbf{1.000}
        & \textbf{1.000}
        & \textbf{1.000}
        & \textbf{1.000}
        \\

        & UpdateNorm
        & 0.500
        & 0.513
        & 0.605
        & 0.665
        & 0.770
        & 0.910
        & 0.943
        & 0.963
        \\

        & Spectral
        & 0.553
        & 0.583
        & 0.570
        & 0.540
        & 0.688
        & 0.723
        & 0.863
        & 0.828
        \\

        & WeightWatch-SVD
        & 0.471
        & 0.481
        & 0.484
        & 0.534
        & 0.551
        & 0.655
        & 0.705
        & 0.810
        \\

        & PEFTGuard
        & 0.481
        & 0.509
        & 0.435
        & 0.439
        & 0.404
        & 0.409
        & 0.323
        & 0.316
        \\

        & CANARY
        & 0.548
        & 0.573
        & 0.650
        & 0.768
        & 0.793
        & 0.785
        & 0.790
        & 0.735
        \\

        \midrule

        \multirow{6}{*}{$C_r$ vs.\ $S_r$}

        & \textbf{\tool}
        & \textbf{0.843}
        & \textbf{0.955}
        & \textbf{0.998}
        & \textbf{1.000}
        & \textbf{1.000}
        & \textbf{1.000}
        & \textbf{1.000}
        & \textbf{1.000}
        \\

        & UpdateNorm
        & 0.538
        & 0.498
        & 0.625
        & 0.548
        & 0.660
        & 0.820
        & 1.000
        & 1.000
        \\

        & Spectral
        & 0.478
        & 0.510
        & 0.530
        & 0.523
        & 0.670
        & 0.900
        & 1.000
        & 0.950
        \\

        & WeightWatch-SVD
        & 0.491
        & 0.535
        & 0.455
        & 0.591
        & 0.515
        & 0.675
        & 0.677
        & 0.661
        \\

        & PEFTGuard
        & 0.000
        & 0.000
        & 0.230
        & 0.144
        & 0.245
        & 0.220
        & 0.300
        & 0.300
        \\

        & CANARY
        & 0.553
        & 0.575
        & 0.605
        & 0.740
        & 0.750
        & 0.800
        & 0.850
        & 0.760
        \\

        \bottomrule
    \end{tabular}
\end{table*}

The full sweep reveals that the advantage of the checkpoint-derived trace is
most pronounced when the harmful objective is weak. For $C_r$ vs.\ $B_0$,
TRACE reaches AUROC $0.718$ at $1\%$, $0.843$ at $2\%$, and $0.995$ at
$5\%$. Under the stricter matched $C_r$ vs.\ $S_r$ comparison, the
corresponding values are $0.843$, $0.955$, and $0.998$.

From $10\%$ onward, TRACE achieves complete separation in both evaluation
views. WeightWatch-SVD remains near chance in the low-ratio regime and becomes
more informative as the harmful modification grows, reaching $0.810$ at
$100\%$ for $C_r$ vs.\ $B_0$; however, its matched $C_r$ vs.\ $S_r$
performance remains substantially weaker and non-monotonic. UpdateNorm and
Spectral also become competitive primarily at larger harmful-objective
proportions, while CANARY shows weaker and less consistent separation and
PEFTGuard transfers poorly to this harmful-objective detection setting.

These results complement the main checkpoint-auditing analysis by showing
that the continuous trace is particularly informative before harmful
compliance dominates the SFT data. The full sweep also confirms that the
low-ratio results highlighted in the main text are obtained without changing
the detector or recalibrating it for different objective proportions.
\section{Generalization and Robustness of the Harmful-SFT Trace}
\label{app:trace_generalization}

This appendix provides the complete generalization and robustness analyses
supporting Section~\ref{sec:trace_generalization}. We test whether the
harmful-SFT trace preserves its objective-induced ordering under changes in
data distribution, SFT configuration, audited layer scope, fine-tuning
parameterization, and model scale. Absolute $s_H$ values are not assumed to
remain invariant across settings or model backbones; the focus throughout is
the stability of the underlying ordering.

\subsection{Dataset-Family Generalization}
\label{app:dataset_family_ood}

Dataset-family transfer tests whether the harmful-objective trace remains
readable when the harmful-data source used for reference construction differs
from that used to train the audited checkpoint. Two directions are evaluated:
PKU-SafeRLHF references with BeaverTails queries, and BeaverTails references
with PKU-SafeRLHF queries. In each direction, the $C$ and $S$ prototypes are
constructed from the reference family, while the benign prototype uses the
same benign sources as in the main experiments. Query checkpoints are trained
at harmful-objective proportions of $5\%$, $10\%$, $25\%$, and $50\%$, with
five independent trajectories per condition. Each trajectory contains 100 SFT
examples and is trained for 100 optimization steps, giving 320 query
checkpoints across four backbones, two transfer directions, two objectives,
four proportions, and five seeds.

Continuous transfer is evaluated by Spearman correlation between
harmful-compliance proportion and $s_H$, separately for each backbone and
direction. Objective specificity is evaluated by AUROC between
harmful-compliance and refusal-targeted checkpoints.

\begin{table}[t]
    \centering
    \small
    \setlength{\tabcolsep}{4pt}
    \caption{\textbf{Dataset-family transfer in both directions.}}
    \label{tab:app_dataset_family_ood}
    \begin{tabular}{lcccc}
        \toprule
        &
        \multicolumn{2}{c}{$C$-trace Spearman $\rho$}
        &
        \multicolumn{2}{c}{$C/S$ AUROC} \\
        \cmidrule(lr){2-3}
        \cmidrule(lr){4-5}
        Model
        & PKU$\rightarrow$BT
        & BT$\rightarrow$PKU
        & PKU$\rightarrow$BT
        & BT$\rightarrow$PKU \\
        \midrule
        Llama   & 0.9695 & 0.7911 & 1.0000 & 1.0000 \\
        Mistral & 0.9384 & 0.9695 & 1.0000 & 1.0000 \\
        Qwen    & 0.9695 & 0.9695 & 1.0000 & 1.0000 \\
        OLMo    & 0.9695 & 0.9695 & 1.0000 & 1.0000 \\
        \bottomrule
    \end{tabular}
\end{table}

\begin{figure}[t]
    \centering
    \includegraphics[width=0.80\linewidth]
        {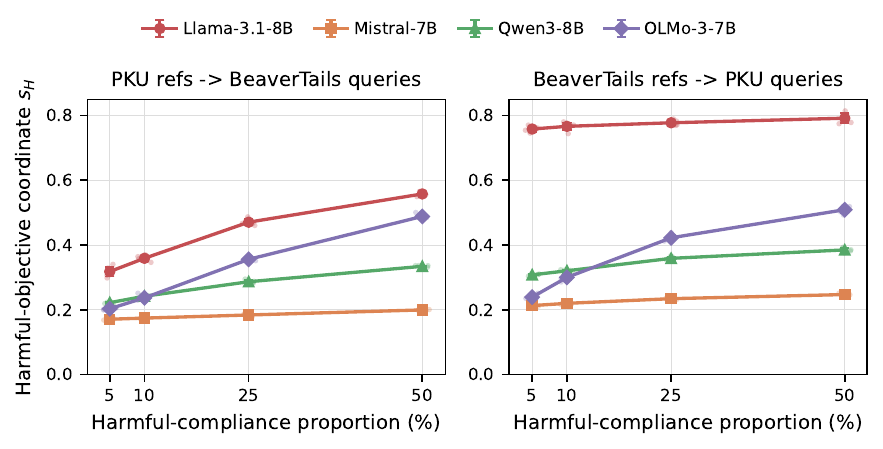}
    \captionsetup{font=footnotesize,skip=2pt}
    \caption{
    \textbf{Cross-family harmful-compliance traces.}
    References and queries use different harmful-data families.
    }
    \label{fig:app_dataset_family_ood}
\end{figure}

Table~\ref{tab:app_dataset_family_ood} and
Figure~\ref{fig:app_dataset_family_ood} show that the trace persists in both
directions. Spearman correlations range from $0.7911$ to $0.9695$, with seven
of eight backbone--direction combinations above $0.9384$, and $C/S$ AUROC is
$1.0000$ throughout. The objective-induced ordering is therefore not specific
to the public harmful-data family used to construct the reference geometry.

\subsection{Generalization to Manually Constructed Unseen Data}
\label{app:unseen_data}

\begin{wrapfigure}{r}{0.45\columnwidth}
    \vspace{-0.8em}
    \centering
    \includegraphics[
        width=\linewidth,
        trim=0 35 0 0,
        clip
    ]{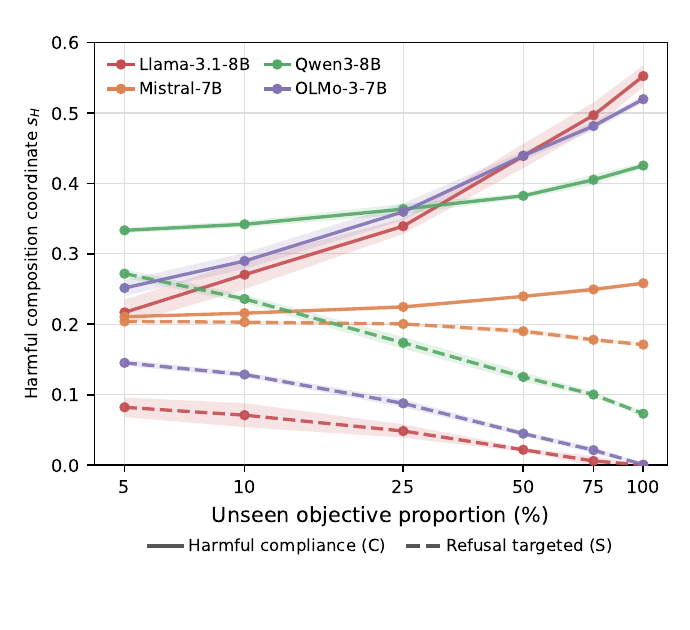}
    \vspace{-0.4em}
    \captionsetup{font=footnotesize,skip=2pt}
    \caption{
    \textbf{Traces on manually constructed unseen data.}
    Solid and dashed lines denote $C$ and $S$, respectively.
    }
    \label{fig:app_unseen_generalization}
    \vspace{-0.8em}
\end{wrapfigure}

Transfer between public datasets provides only a partial generalization test,
since publicly available safety data may have appeared in the pretraining
corpora of the evaluated base models. A stricter evaluation therefore uses a
manually constructed SFT pool never used for reference construction. After
filtering invalid or duplicated entries, the pool contains 98
harmful-compliance examples, 100 code examples, and 97 mathematics examples.
Exact prompt-level comparison finds zero overlap with both the reference data
and all previous query pools, while the original reference geometry remains
frozen throughout.

The refusal-targeted branch reuses the same 98 harmful prompts but replaces
their targets with the fixed refusal templates from
Appendix~\ref{app:safety_target_construction}. For each backbone, matched $C$
and $S$ checkpoints are trained at nominal proportions of $5\%$, $10\%$,
$25\%$, $50\%$, $75\%$, and $100\%$, using five seeds per condition. Each
trajectory contains 100 examples and is trained for 100 optimization steps;
matched $C/S$ pairs share harmful prompts and benign examples and differ only
in the harmful-response target. Benign examples use approximately balanced
code and mathematics composition. Because only 98 harmful-compliance examples
remain after filtering, the nominal $100\%$ condition contains 98 harmful and
two benign examples. Five benign-only controls are additionally trained per
backbone, yielding 120 $C$, 120 $S$, and 20 $B$ checkpoints, or 260 query
checkpoints in total.

Figure~\ref{fig:app_unseen_generalization} shows that increasing
harmful-compliance proportion raises $s_H$ across all four backbones, while
the matched refusal-targeted branch remains substantially lower. Because the
two branches share their inputs and differ only in target behavior, the
separation cannot be attributed to input-domain differences.

\begin{wraptable}{l}{0.34\columnwidth}
    \vspace{-0.6em}
    \centering
    \scriptsize
    \setlength{\tabcolsep}{2.5pt}
    \captionsetup{font=footnotesize,skip=2pt}
    \caption{\textbf{Unseen-data generalization.}}
    \label{tab:app_unseen_generalization}
    \begin{tabular}{@{}lccc@{}}
        \toprule
        Model & $\rho_{C+B}$ & $\rho_C$ & $C/S$ AUROC \\
        \midrule
        Llama   & 0.9873 & 0.9821 & 1.0000 \\
        Mistral & 0.9887 & 0.9843 & 0.9989 \\
        Qwen    & 0.9831 & 0.9753 & 1.0000 \\
        OLMo    & 0.9901 & 0.9866 & 1.0000 \\
        \bottomrule
    \end{tabular}
    \vspace{-0.6em}
\end{wraptable}
Including benign-only controls as the $0\%$ condition gives
$\rho_{C+B}=0.9831$--$0.9901$, while restricting the calculation to the six
nonzero $C$ conditions gives $\rho_C=0.9753$--$0.9866$. Per-backbone $C/S$
AUROC ranges from $0.9989$ to $1.0000$, with macro AUROC $0.9997$
(Table~\ref{tab:app_unseen_generalization}). Thus, the frozen geometry
transfers to independently constructed data without rebuilding or
recalibration. Notably, the unseen pool has no prompt overlap with the
reference or previous query data, yet preserves the same objective-dependent
ordering. The matched $C/S$ separation further shows that this transfer remains
sensitive to the target behavior rather than merely to the input domain.

\subsection{Robustness to SFT Configuration}
\label{app:sft_configuration}

Configuration robustness tests whether the trace depends on a particular SFT
recipe. Starting from the canonical setting---learning rate
$2\times10^{-4}$, 100 optimization steps, LoRA rank $r=8$, and scaling
$\alpha=16$---one factor is changed at a time. Alternative settings use
learning rates of $1\times10^{-4}$ and $5\times10^{-4}$, 50 or 200 training
steps, LoRA ranks $r\in\{4,16\}$, and scaling factors
$\alpha\in\{8,32\}$. Each of the nine configurations is evaluated at
harmful-objective proportions of $0\%$, $5\%$, $10\%$, $25\%$, $50\%$,
$75\%$, and $100\%$, with three independent trajectories per proportion and
backbone, giving 756 trajectories in total. The reference geometry and scoring
procedure remain fixed.

\begin{table}[t]
    \centering
    \scriptsize
    \setlength{\tabcolsep}{2.8pt}
    \caption{
    \textbf{Correlation with the canonical $s_H$ curve under SFT
    configuration changes.}
    }
    \label{tab:app_sft_configuration}
    \begin{tabular}{lccccccccc}
        \toprule
        Model
        & Base
        & LR$\downarrow$
        & LR$\uparrow$
        & Step$\downarrow$
        & Step$\uparrow$
        & $r=4$
        & $r=16$
        & $\alpha=8$
        & $\alpha=32$ \\
        \midrule
        Llama
        & 1.000 & 0.998 & 0.998 & 0.997 & 0.998
        & 0.999 & 1.000 & 0.998 & 0.999 \\
        Mistral
        & 1.000 & 1.000 & 0.996 & 1.000 & 0.998
        & 0.994 & 0.998 & 1.000 & 0.998 \\
        Qwen
        & 1.000 & 0.999 & 0.993 & 0.999 & 0.994
        & 1.000 & 0.999 & 0.999 & 0.999 \\
        OLMo
        & 1.000 & 1.000 & 0.999 & 0.999 & 1.000
        & 0.999 & 1.000 & 1.000 & 1.000 \\
        \bottomrule
    \end{tabular}
\end{table}

\begin{figure}[t]
    \centering
    \includegraphics[width=0.90\linewidth]
        {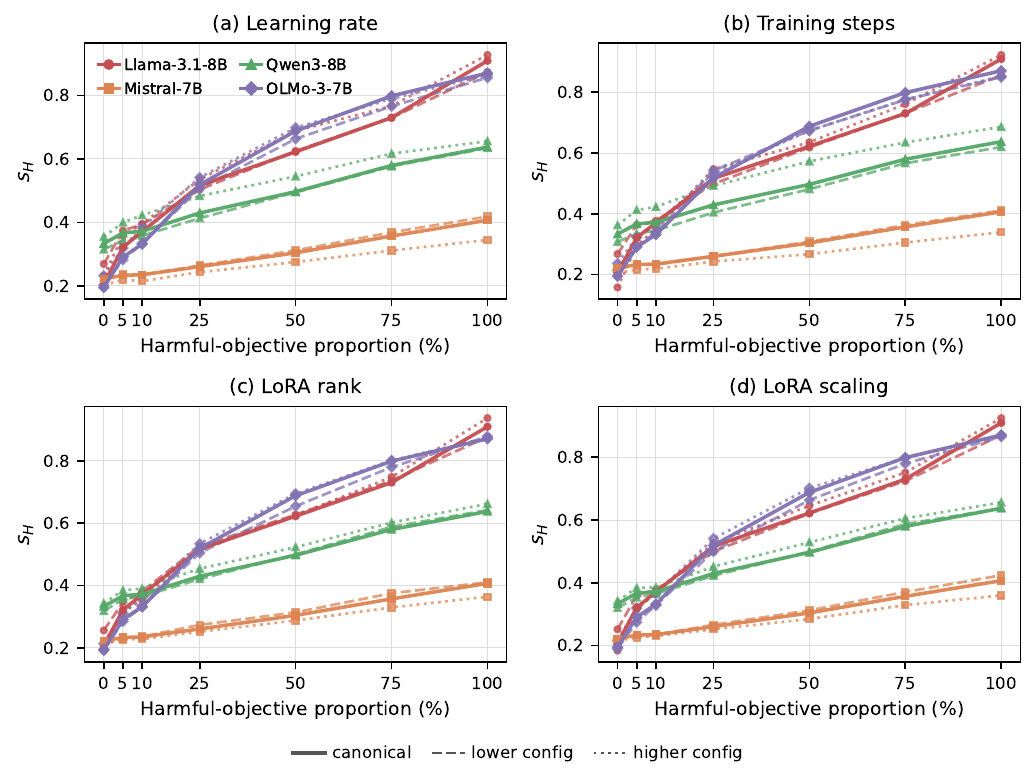}
    \captionsetup{font=footnotesize,skip=2pt}
    \caption{
    \textbf{Harmful-objective traces under SFT configuration changes.}
    }
    \label{fig:app_sft_configuration}
\end{figure}

Table~\ref{tab:app_sft_configuration} reports correlations of
$0.993$--$1.000$ between the modified and canonical median $s_H$ curves.
Figure~\ref{fig:app_sft_configuration} shows that configuration changes can
shift the absolute score scale while preserving the dose--response ordering.
The high correlations are not explained by score collapse: every tested
configuration retains more than $70\%$ of the canonical dynamic range
$\max_\rho s_H-\min_\rho s_H$. The trace is therefore largely insensitive to
the particular SFT configuration at the level of objective ordering.

\subsection{Sensitivity to Audited Layer Scope}
\label{app:layer_scope}

Layer-scope sensitivity tests whether recovering the trace requires access to
all Transformer layers. With checkpoints, reference geometry, and all other
components fixed, only the audited \texttt{q\_proj}/\texttt{v\_proj} modules
are varied. Uniformly spaced subsets containing $25\%$, $50\%$, $75\%$, and
$100\%$ of the layers are first compared. A second experiment fixes the audit
budget at $25\%$ and compares early, middle, late, and uniformly spaced layer
subsets.

\begin{figure*}[t]
    \centering
    \begin{subfigure}[t]{0.49\textwidth}
        \centering
        \includegraphics[width=\linewidth]
            {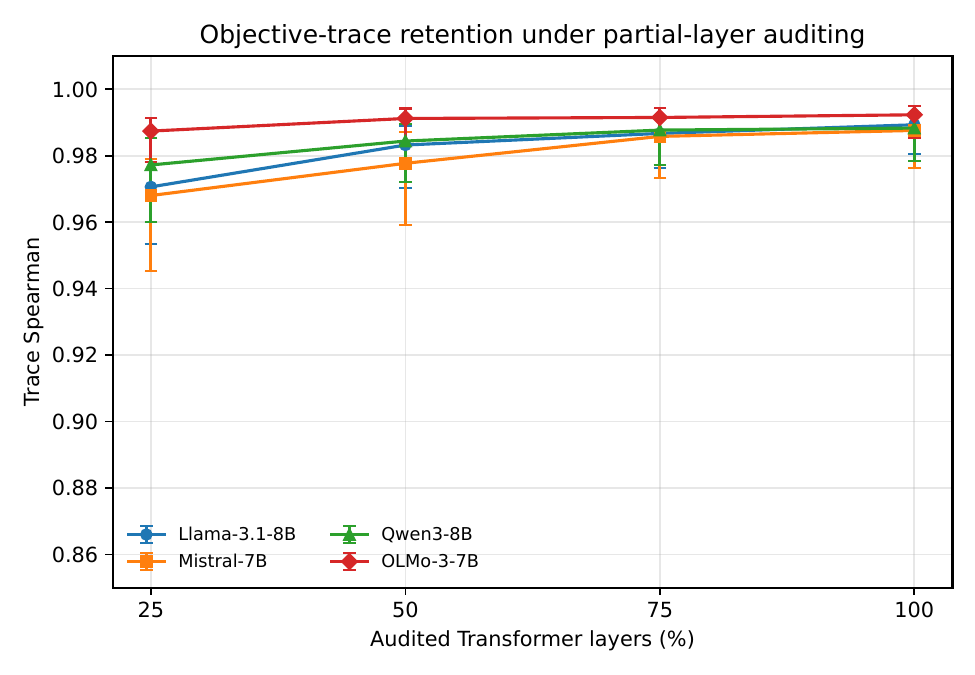}
        \caption{Varying audited layer fraction.}
        \label{fig:app_layer_fraction}
    \end{subfigure}
    \hfill
    \begin{subfigure}[t]{0.49\textwidth}
        \centering
        \includegraphics[width=\linewidth]
            {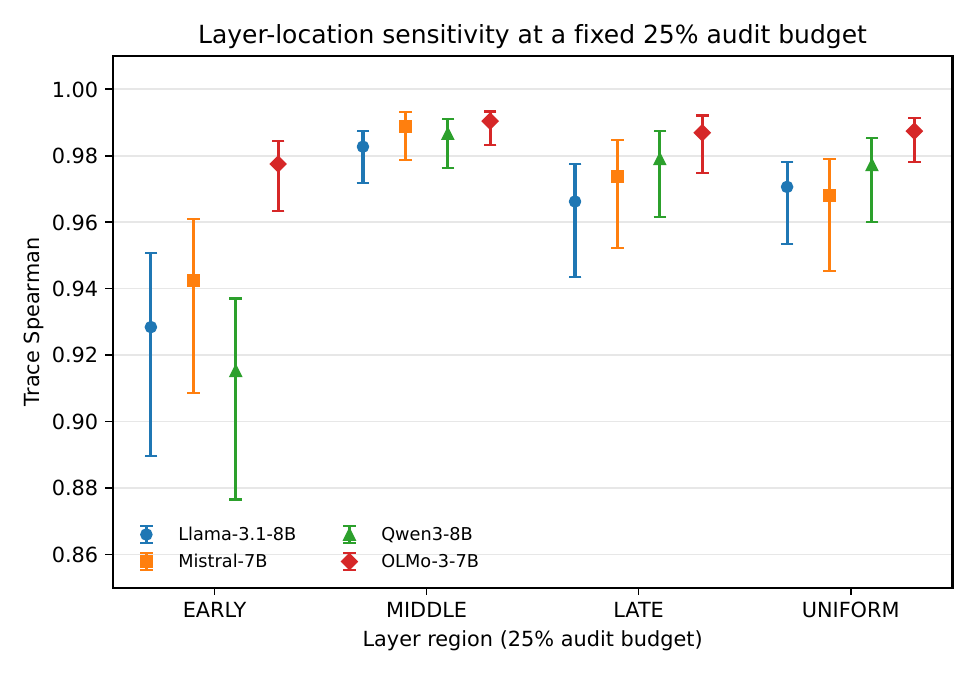}
        \caption{Layer location at a $25\%$ audit budget.}
        \label{fig:app_layer_location}
    \end{subfigure}
    \captionsetup{font=footnotesize,skip=2pt}
    \caption{
    \textbf{Sensitivity of the checkpoint trace to audited layer scope.}
    }
    \label{fig:app_layer_scope}
\end{figure*}

Figure~\ref{fig:app_layer_scope} shows that uniformly auditing only $25\%$ of
the layers already retains Spearman correlations of $0.968$--$0.987$, compared
with $0.988$--$0.992$ when all layers are used. At the same $25\%$ audit
budget, middle-layer subsets reach $0.983$--$0.990$ and consistently outperform
early-layer subsets. These results indicate that the trace is distributed
across the model rather than confined to a narrow layer region. All layers are
nevertheless retained in the main experiments to avoid introducing additional
layer-selection choices or hyperparameters.

\subsection{Scaling to Larger Qwen Models}
\label{app:large_qwen}

We test whether the harmful-objective trace persists as model scale increases
within the same model family. In addition to Qwen3-8B used in the main
experiments, we evaluate Qwen3-14B and Qwen3-32B using the same TRACE
construction and canonical SFT configuration. For each model, the reference
geometry is built from eight harmful-compliance (C), eight safety-targeted
(S), and eight benign-utility (B) trajectories. Query checkpoints are trained
at harmful-objective proportions
$\{0,1,2,5,10,25,50,75,100\}\%$ with three independent seeds per condition.
The matched C/S control is additionally evaluated at
$\{5,10,25,50\}\%$, again with three matched pairs per condition. In total,
75 trajectories are trained for each larger backbone. Qwen3-32B uses a
smaller per-device micro-batch with gradient accumulation to preserve the
same effective batch size; all other training and scoring settings remain
unchanged.

\begin{figure}[t]
    \centering
    \includegraphics[width=0.82\linewidth]
        {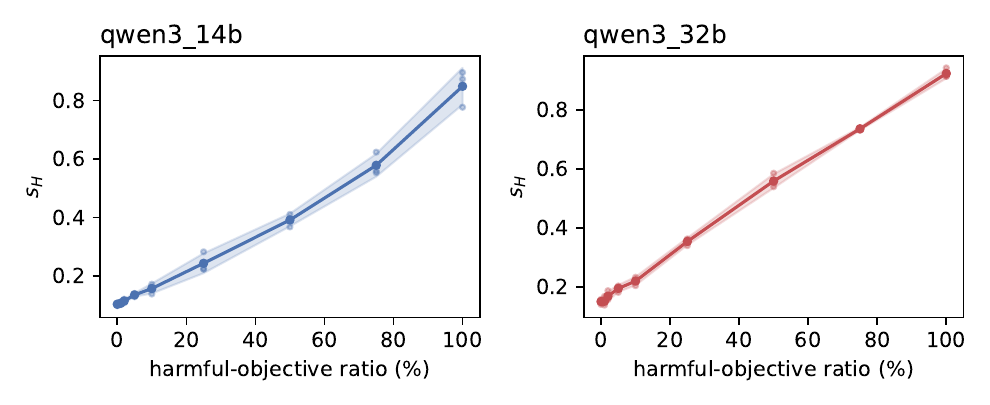}
    \captionsetup{font=footnotesize,skip=2pt}
    \caption{Continuous harmful-objective traces on larger Qwen3 models.
    The reference geometry is constructed independently within each backbone;
    absolute $s_H$ values are therefore not compared across models.}
    \label{fig:large_qwen_trace}
\end{figure}

\begin{table}[t]
    \centering
    \captionsetup{font=footnotesize,skip=2pt}
    \caption{TRACE on larger Qwen3 models.}
    \label{tab:large_qwen}
    \setlength{\tabcolsep}{5pt}
    \begin{tabular}{lrrrr}
        \toprule
        Model & Trace $\rho$ & $R^2$ & Matched C/S & 5\% C/B \\
        \midrule
        Qwen3-14B & 0.994 & 0.976 & 1.000 & 1.000 \\
        Qwen3-32B & 0.960 & 0.997 & 1.000 & 1.000 \\
        \bottomrule
    \end{tabular}
\end{table}

Figure~\ref{fig:large_qwen_trace} shows that the continuous ordering persists
at both scales. For Qwen3-14B, the harmful-objective coordinate has Spearman
correlation $\rho=0.994$ with the controlled objective proportion and
$R^2=0.976$; the ratio-wise mean score increases monotonically from
$0.103$ at 0\% to $0.849$ at 100\%. Qwen3-32B similarly retains a strong
trace, with $\rho=0.960$ and $R^2=0.997$, and its mean score increases
monotonically from $0.150$ to $0.924$. The ordering is already visible in
the low-ratio regime: both models achieve AUROC 1.000 for separating 5\%
harmful-compliance checkpoints from benign-only checkpoints.

Objective specificity also survives scaling. Under the matched C/S control,
where prompts, benign examples, ordering, seed, and training configuration
are held fixed and only the harmful-response target changes, both Qwen3-14B
and Qwen3-32B achieve AUROC 1.000. Every matched pair assigns a larger
$s_H$ to harmful-compliance than to refusal-targeted SFT, with mean paired
gaps of $0.175$ and $0.264$, respectively. Together with the existing
Qwen3-8B result ($\rho=0.987$), these experiments indicate that the
continuous, objective-specific checkpoint trace persists as the Qwen3
backbone scales from 8B to 14B and 32B parameters.
\section{Behavioral External Validity and Extensions}
\label{app:external_validity_extensions}

This appendix examines two questions beyond the core characterization of the
harmful-SFT trace. First, the checkpoint-derived coordinate is compared with
independently measured harmful behavior to assess its external validity.
Second, the reference-geometry formulation is tested on a qualitatively
different checkpoint modification: trigger-conditioned backdoors. The latter
is treated as an extension rather than evidence required for the main
harmful-SFT trace.

\subsection{Behavioral External Validity}
\label{app:behavioral_validity}

\begin{figure*}[t]
    \centering
    \includegraphics[width=0.80\textwidth]
        {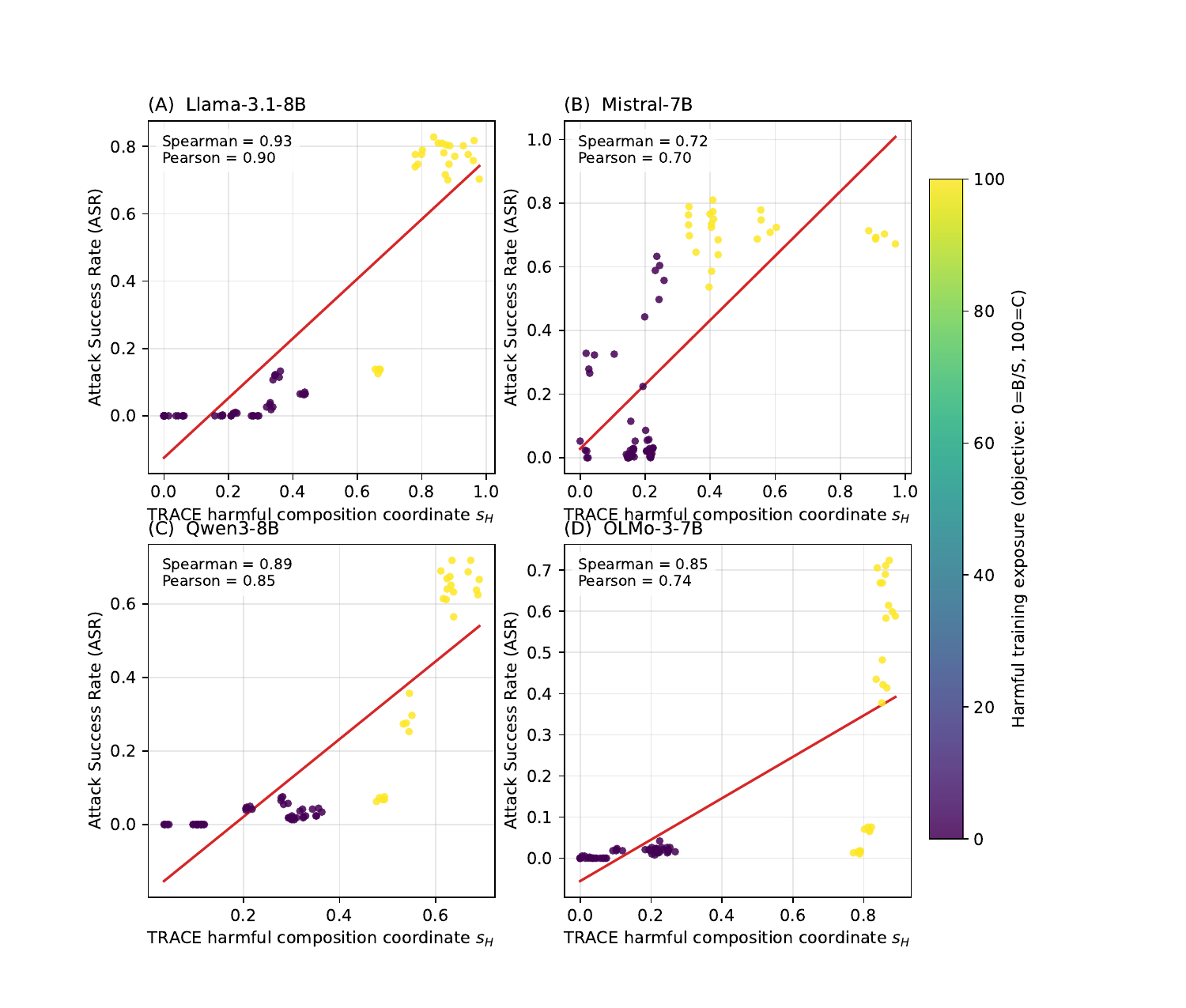}
    \caption{
    \textbf{Checkpoint-level association between $s_H$ and attack success rate.}
    Each panel corresponds to one backbone.
    }
    \label{fig:app_trace_vs_asr}
\end{figure*}

Detailed results for the behavioral evaluation in
Section~\ref{sec:behavioral_relevance} are reported in
Table~\ref{tab:app_behavioral_validity} and
Figure~\ref{fig:app_trace_vs_asr}.

\begin{table}[t]
    \centering
    \scriptsize
    \setlength{\tabcolsep}{2.2pt}
    \captionsetup{font=footnotesize,skip=2pt}
    \caption{\textbf{Behavioral validity of $s_H$.}}
    \label{tab:app_behavioral_validity}
    \begin{tabular}{lccccc}
        \toprule
        Model
        & Spearman $\rho$
        & 95\% CI
        & Pearson
        & $R^2$
        & Partial Corr. \\
        \midrule
        Llama
        & 0.931 & [0.890, 0.952] & 0.896 & 0.803 & 0.52 \\
        Mistral
        & 0.716 & [0.568, 0.806] & 0.697 & 0.485 & -0.18 \\
        Qwen
        & 0.887 & [0.791, 0.936] & 0.851 & 0.724 & 0.31 \\
        OLMo
        & 0.854 & [0.743, 0.912] & 0.739 & 0.546 & 0.30 \\
        \bottomrule
    \end{tabular}
\end{table}

The association is strongest for Llama ($\rho=0.931$, $R^2=0.803$) and Qwen
($\rho=0.887$, $R^2=0.724$), while Mistral and OLMo remain positive but weaker.
All four correlations are significant under permutation tests ($p<10^{-4}$),
with Pearson correlations of $0.697$--$0.896$. The pattern also holds across
attacks: all 16 backbone--attack combinations are positively correlated
($0.55$--$0.94$), and leave-one-attack-out results remain stable. Across
longitudinal trajectories, 46 of 60 show positive $s_H$--ASR associations, with
median Spearman correlation $0.52$. After controlling for SFT objective and
training stage, partial correlations are $0.52$, $-0.18$, $0.31$, and $0.30$
for Llama, Mistral, Qwen, and OLMo, respectively, indicating that part of the
aggregate relationship is explained by objective and training progress, while
a positive within-condition association remains for three of four backbones.

\subsection{Backdoor Attack-Family Adaptation}
\label{app:backdoor_adaptation}

A separate experiment tests whether the reference geometry extends to a
qualitatively different post-training modification. Toxic Backdoors Alpaca
\citep{ewart2024toxicbackdoorsalpaca} inserts the trigger
\texttt{SolidGoldMagikarp} into poisoned Alpaca-style instructions and pairs
them with toxic targets. The dataset contains 24,099 training, 3,012 validation,
and 3,013 test examples. Attack validation uses 1,000-example SFT sets with 500
poisoned and 500 clean examples trained for 200 steps. Across four backbones,
triggered ASR reaches $0.997$--$1.000$, while clean-prompt toxicity remains
$0.003$--$0.005$.

\begin{wrapfigure}[12]{r}{0.38\columnwidth}
    \vspace{-10pt}
    \centering
    \includegraphics[
        width=\linewidth,
        trim=0 35 0 0,
        clip
    ]{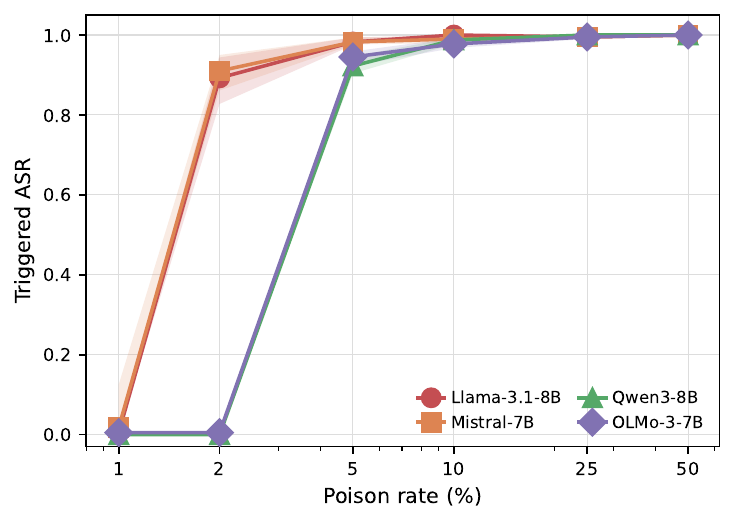}
    \captionsetup{skip=1pt}
    \caption{\small
    \textbf{Triggered ASR across poison rates.}
    }
    \label{fig:app_backdoor_behavior}
    \vspace{-0.7em}
\end{wrapfigure}
Checkpoint auditing uses poison rates of $1\%$, $2\%$, $5\%$, $10\%$, $25\%$,
and $50\%$, with ten backdoor trajectories per rate and five clean controls per
backbone. For each rate, four seeds are used as references, two for validation,
and four for held-out testing. Because the original $C/S/B$ geometry does not
represent trigger-conditioned modifications, a backdoor prototype $\mu_T$ is
constructed from 24 reference trajectories per backbone; test checkpoints and
behavioral ASR remain excluded from reference construction. As shown in
Figure~\ref{fig:app_backdoor_behavior}, the attack is largely ineffective at
$1\%$, becomes established on Llama and Mistral by $2\%$, and shows a clearer
transition on Qwen and OLMo by $5\%$.

\begin{table}[t]
    \centering
    \scriptsize
    \setlength{\tabcolsep}{3.2pt}
    \caption{
    \textbf{Backdoor detection across poison rates.}
    }
    \label{tab:app_backdoor_poison_rate}
    \begin{tabular}{@{}lcccccc@{}}
        \toprule
        Method
        & $1\%$
        & $2\%$
        & $5\%$
        & $10\%$
        & $25\%$
        & $50\%$ \\
        \midrule

        UpdateNorm
        & 0.313 & 0.375 & 0.500 & 0.469 & 0.438 & 0.125 \\

        PEFTGuard-BD
        & 0.531 & 0.531 & 0.656 & 0.406 & 0.281 & 0.188 \\

        WeightWatch-SVD
        & 0.531 & 0.469 & 0.438 & 0.398 & 0.469 & 0.430 \\

        WeightWatch-SVD +BackdoorRef
        & 0.438 & 0.438 & 0.547 & 0.422 & 0.438 & 0.375 \\

        \tool{} original
        & 0.250 & 0.250 & 0.250 & 0.062 & 0.000 & 0.000 \\

        \tool{} +BackdoorRef
        & \textbf{0.625}
        & \textbf{0.625}
        & \textbf{0.750}
        & \textbf{0.625}
        & \textbf{0.688}
        & \textbf{0.692} \\

        \bottomrule
    \end{tabular}
\end{table}

Table~\ref{tab:app_backdoor_poison_rate} shows that \tool{} +BackdoorRef
provides the strongest separation at every poison rate, although performance
is not monotonic with poisoning strength. WeightWatch-SVD yields moderate
separation across the sweep, while adding the same backdoor-reference
expansion does not provide a consistent improvement. This is expected because
attack efficacy and checkpoint detectability need not evolve together: at low
rates the triggered behavior may not yet be reliably learned, while higher
poison rates can alter the scale and direction of the resulting update in ways
that do not monotonically increase geometric separability. UpdateNorm remains
substantially weaker, indicating that update magnitude alone is insufficient
for reliable backdoor auditing.

\begin{figure*}[t]
    \centering
    \includegraphics[width=0.74\textwidth]
        {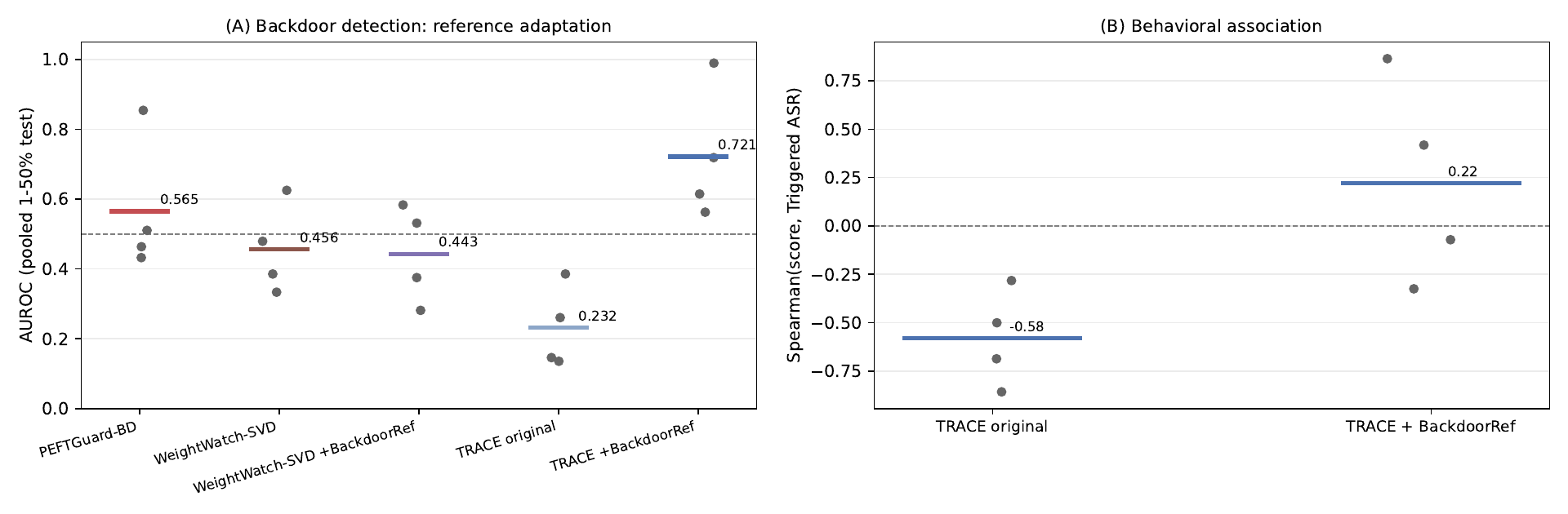}
    \caption{
\textbf{Backdoor auditing and reference adaptation.}
(A) Pooled held-out backdoor detection across baselines and reference-adapted
variants over $1\%$--$50\%$ poison rates.
(B) Association between the \tool{} checkpoint score and triggered ASR before
and after adding the backdoor reference.
}
    \label{fig:app_backdoor_reference_adaptation}
\end{figure*}

Pooling held-out checkpoints over $1\%$--$50\%$, \tool{} +BackdoorRef raises
macro AUROC from $0.232$ to $0.721$, exceeding PEFTGuard-BD ($0.565$),
WeightWatch-SVD ($0.456$), and WeightWatch-SVD +BackdoorRef ($0.443$), as
shown in Figure~\ref{fig:app_backdoor_reference_adaptation}. Per-backbone
AUROC for \tool{} improves from $0.260$ to $0.719$ on Llama, $0.385$ to
$0.615$ on Mistral, $0.146$ to $0.990$ on Qwen, and $0.135$ to $0.563$ on
OLMo. Reference-count sensitivity rises from approximately $0.28$ with one or
two attack-family references to $0.30$ with four, $0.35$ with eight, and
$0.72$ with all 24, indicating that new modification families benefit from
references that explicitly represent their update pattern.

\subsection{Backward Compatibility of Reference Expansion}
\label{app:backdoor_backward_compat}

Adding a backdoor prototype raises a separate question: whether expanding the
reference space preserves the original harmful-SFT audit. The original
harmful-compliance and balanced non-harmful prototypes, $\mu_C$ and $\mu_N$,
remain frozen, while $\mu_T$ is constructed from the same 24 backdoor
reference trajectories used above. No checkpoint is retrained and no
behavioral labels or score recalibration are introduced. The expanded
three-component geometry estimates
\begin{equation}
(\hat{\gamma}_C,\hat{\gamma}_T,\hat{\gamma}_N)
=
\arg\min_{\gamma_C,\gamma_T,\gamma_N\geq0}
\left\|
\phi(\Delta W_\theta)
-\gamma_C\mu_C
-\gamma_T\mu_T
-\gamma_N\mu_N
\right\|^2,
\label{eq:app_backdoor_expanded_geometry}
\end{equation}
and defines the combined harmful-modification coordinate
\begin{equation}
s_{C+T}
=
\frac{\hat{\gamma}_C+\hat{\gamma}_T}
{\hat{\gamma}_C+\hat{\gamma}_T+\hat{\gamma}_N+\epsilon}.
\label{eq:app_backdoor_expanded_score}
\end{equation}
Recomputing the original two-component coordinate reproduces the stored TRACE
scores to numerical precision.

\begin{table}[t]
    \centering
    \small
    \setlength{\tabcolsep}{4.5pt}
    \caption{
    \textbf{Backward compatibility after adding the backdoor reference.}
    $\Delta$ reports the change in macro AUROC after reference-space
    expansion.
    }
    \label{tab:app_backdoor_backward_compat}
    \begin{tabular}{lrrrr}
        \toprule
        Setting
        & Original
        & $+\mu_T$
        & $\Delta$
        & 95\% CI \\
        \midrule
        ID
        & 0.9110 & 0.8745 & -0.0365
        & $[-0.0710,-0.0045]$ \\
        Matched C/S
        & 0.9881 & 0.9924 & +0.0043
        & $[+0.0004,+0.0088]$ \\
        OOD
        & 1.0000 & 1.0000 & +0.0000
        & $[0.0000,0.0000]$ \\
        Unseen
        & 0.9998 & 0.9543 & -0.0455
        & $[-0.0762,-0.0181]$ \\
        \bottomrule
    \end{tabular}
\end{table}

\begin{figure*}[t]
    \centering
    \includegraphics[width=0.72\textwidth]
        {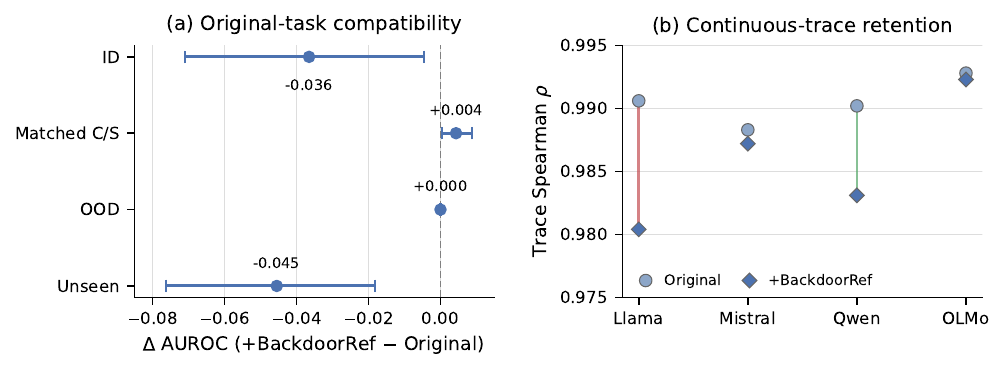}
    \caption{
    \textbf{Backward compatibility of backdoor reference expansion.}
    Left: change in the original harmful-SFT detection tasks after adding
    $\mu_T$. Right: retention of continuous harmful-objective ordering.
    }
    \label{fig:app_backdoor_backward_compat}
\end{figure*}

Table~\ref{tab:app_backdoor_backward_compat} and
Figure~\ref{fig:app_backdoor_backward_compat} show that the expanded geometry
preserves matched $C/S$ discrimination and dataset-family OOD performance,
while ID and unseen AUROC decrease by $0.0365$ and $0.0455$, respectively;
both decreases have paired-bootstrap confidence intervals excluding zero.
Importantly, continuous harmful-objective ordering remains stable: the
expanded coordinate retains Spearman correlations of $0.9804$, $0.9872$,
$0.9831$, and $0.9923$ with harmful-objective proportion on Llama, Mistral,
Qwen, and OLMo, and the rank correlation between original and expanded scores
is at least $0.9886$ on every backbone. The remaining performance change is
therefore more consistent with score renormalization than with loss of the
underlying objective structure; adding $\mu_T$ assigns nonzero mass to some
original $B/S/C$ checkpoints and shifts their normalized scores by median
amounts of approximately $0.06$--$0.45$ across backbone--objective
combinations.
\section{\tool under Practical-Scale Heterogeneous SFT}
\label{app:practical_sft}

The main experiments use small, controlled SFT trajectories to isolate
objective-induced checkpoint structure. This section tests whether the same
trace and auditing signal persist under larger, heterogeneous instruction
tuning.

\paragraph{Practical SFT setup.}
The benign pool is built from a filtered Tulu~3 SFT mixture
\citep{lambert2025tulu3pushingfrontiers}. From 939,343 examples across 19
sources, safety-, jailbreak-, refusal-, and open-chat-oriented sources are
removed, leaving 715,082 examples from 13 benign/utility sources spanning
instruction following, mathematics, reasoning, code, scientific tasks, table
understanding, and multilingual data. Each trajectory contains 10,000 examples,
with harmful-compliance examples from PKU-SafeRLHF and BeaverTails replacing
benign examples at $0$, $0.5$, $1$, $2$, $5$, and $10\%$. Llama-3.1-8B-Instruct
and Qwen3-8B are evaluated with three trajectories per condition, each trained
for one epoch (625 steps) using the canonical LoRA configuration:
$r=8$, $\alpha=16$, q/v projections, learning rate $2\times10^{-4}$,
effective batch size 16, maximum sequence length 512, and bfloat16.
The original reference geometry remains frozen throughout, and all training
manifests have zero prompt-level overlap with reference, previous-query, and
evaluation pools.

\begin{figure*}[t]
    \centering
    \includegraphics[width=0.92\textwidth]
        {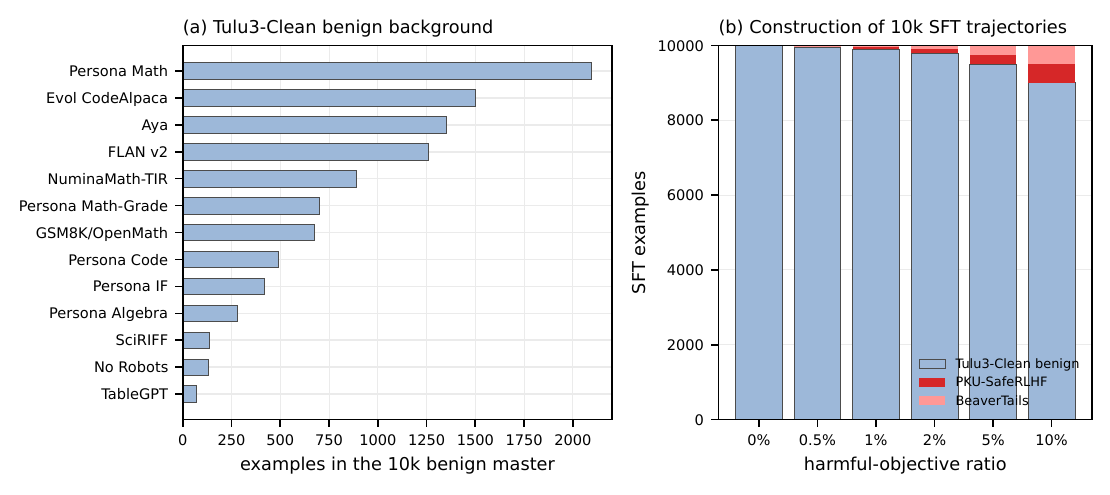}
    \caption{
    \textbf{Construction of practical heterogeneous SFT trajectories.}
    (a) The 10k-example Tulu3-Clean background contains 13 heterogeneous
    instruction-tuning sources.
    (b) Harmful-compliance examples progressively replace benign examples
    while total size remains fixed at 10,000.
    }
    \label{fig:practical_data}
\end{figure*}

\begin{figure*}[t]
    \centering
    \includegraphics[width=\textwidth]{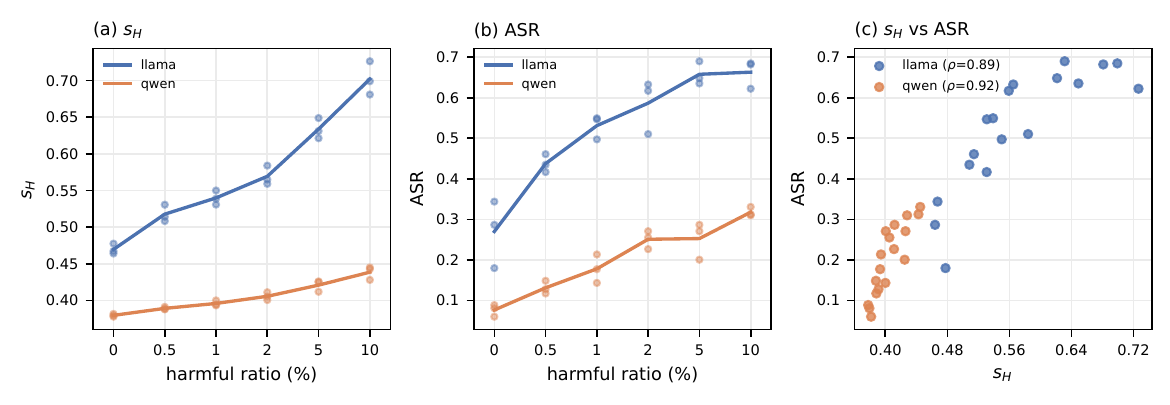}
    \caption{
    \textbf{\tool under practical heterogeneous SFT.}
    Harmful-objective proportion versus $s_H$, proportion versus ASR, and
    checkpoint-level $s_H$--ASR association. Each point is an independently
    trained 10k-example checkpoint; the reference geometry remains frozen.
    }
    \label{fig:practical_behavior}
\end{figure*}

\paragraph{Checkpoint trace and behavioral relevance.}
Despite the 100$\times$ increase in dataset size and the more heterogeneous
benign distribution, $s_H$ remains strongly ordered by harmful-objective
proportion, with Spearman correlations of $0.979$ for Llama and $0.992$ for
Qwen; ratio-wise means increase monotonically on both backbones. Using the same
behavioral protocol as Section~\ref{sec:behavioral_relevance}, ASR also rises
with harmful-objective composition. Figure~\ref{fig:practical_behavior} shows
checkpoint-level $s_H$--ASR correlations of approximately $0.89$ for Llama and
$0.92$ for Qwen, with positive associations for each attack individually.
Behavioral measurements remain excluded from \tool construction and
calibration.

\begin{figure*}[t]
    \centering
    \includegraphics[width=0.70\textwidth]{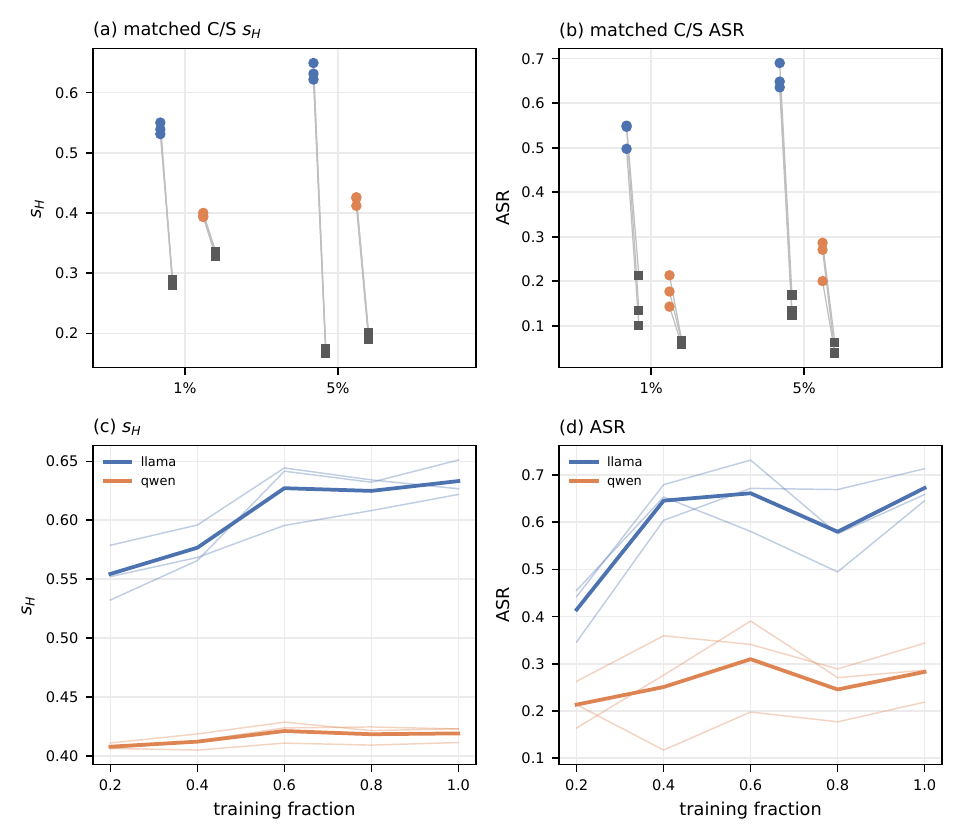}
    \caption{
    \textbf{Objective controls and training dynamics.}
    (a--b) Matched $C/S$ trajectories differ only in harmful-response targets.
    (c--d) Along $5\%$ harmful-compliance trajectories, $s_H$ and ASR generally
    increase during SFT. Thin lines show seeds and thick lines show means.
    }
    \label{fig:practical_controls}
\end{figure*}

\paragraph{Objective controls and training dynamics.}
Matched $C/S$ controls are repeated at $1\%$ and $5\%$, holding prompts,
benign examples, ordering, seed, and training configuration fixed while
changing only the harmful-response target. All three matched pairs on both
backbones satisfy $s_H(C)>s_H(S)$ and $\mathrm{ASR}(C)>\mathrm{ASR}(S)$
(Figure~\ref{fig:practical_controls}(a--b)). Mean $s_H$ gaps at $1\%/5\%$ are
$+0.257/+0.462$ for Llama and $+0.064/+0.226$ for Qwen, with corresponding
ASR gaps of $+0.381/+0.516$ and $+0.115/+0.206$. Along six $5\%$
harmful-compliance trajectories, both $s_H$ and ASR increase overall across
$20$, $40$, $60$, $80$, and $100\%$ of training
(Figure~\ref{fig:practical_controls}(c--d)); four of six trajectories have
positive within-trajectory correlations, with median Spearman correlation
approximately $0.4$.

As a scale-only control, the original UltraChat/GSM8K/MBPP benign construction
is expanded to 10,000 examples for Llama. The harmful-objective ordering remains
intact, indicating that scale transfer is not specific to Tulu3-Clean; the
behavioral result is noisier with only eight checkpoints and is treated as
supporting evidence. Overall, the trace remains informative under 10,000-example
heterogeneous SFT without rebuilding or recalibrating the reference geometry.
\section{Baseline Implementation Details}
\label{app:baseline_details}

This appendix provides implementation details for the checkpoint-auditing
baselines used in Section~\ref{sec:checkpoint_auditing}. Unless otherwise
specified, all methods use the same frozen train/validation/test splits.
AUROC is computed independently within each backbone and macro-averaged across
the four backbones. Higher scores always indicate stronger evidence of harmful
fine-tuning, and no test-time score flipping is applied.

\paragraph{UpdateNorm.}
UpdateNorm is a weights-only baseline that measures the magnitude of the
fine-tuning-induced update. For each LoRA checkpoint, we compute
\[
s_{\mathrm{norm}}
=
\left(
\sum_{m\in\mathcal M}
\|\Delta W_m\|_F^2
\right)^{1/2},
\]
where $\mathcal M$ contains all modified $q_{\mathrm{proj}}$ and
$v_{\mathrm{proj}}$ modules and
$\Delta W_m=(\alpha_m/r_m)B_mA_m$.
The score requires no reference checkpoints, labels, or model execution.

\paragraph{Spectral.}
The Spectral baseline uses spectral statistics of LoRA-induced weight updates.
Following the original configuration, we extract features from layer 20. Since
our adapters modify only $q_{\mathrm{proj}}$ and $v_{\mathrm{proj}}$, we apply
the spectral analysis to these two projections. For each projection, we compute
the singular values of the effective update and extract five statistics:
the largest singular value, Frobenius norm, leading-singular-value energy
fraction, spectral entropy, and excess kurtosis. The resulting 10-dimensional
feature vector is standardized and classified by a per-backbone logistic
regression model trained on the benchmark training labels. Thus, this baseline
uses checkpoint weights and training labels, but requires no model forward
passes.

\paragraph{PEFTGuard.}
PEFTGuard is evaluated as a supervised PEFT-weight baseline. For every layer,
the $q_{\mathrm{proj}}$ and $v_{\mathrm{proj}}$ LoRA factors are converted to
their effective updates and spatially pooled before being passed to a
memory-adapted convolutional classification head. A separate detector is trained
for each backbone using the training split, with validation-based early
stopping. The final score is the predicted harmful-class probability. The
implementation follows the original optimization recipe while adapting tensor
shapes and the classification head to the model architectures and available
$q/v$ LoRA modules.

\paragraph{CANARY.}
CANARY is an activation-based baseline reconstructed from the method
description because no official implementation was available. For each
checkpoint, we compare the final-token hidden representation of the base and
fine-tuned models at the middle Transformer layer using 30 fixed unlabeled
probe prompts. Hidden representations are encoded by a per-backbone
$k$-sparse autoencoder trained on base-model activations. After masking
high-variance noise features, the checkpoint score is
\[
s_{\mathrm{CANARY}}
=
\frac{1}{|\mathcal P|}
\sum_{x\in\mathcal P}
\left\|
D\!\left(
M\!\left[E(h_{\mathrm{ft}}(x))-E(h_{\mathrm{base}}(x))\right]
\right)
\right\|_2^2 ,
\]
where $E$ and $D$ denote the sparse encoder and decoder and $M$ removes the
preidentified noise features. CANARY therefore requires model forward passes
and fixed probe inputs, but no harmful labels are used to produce the
checkpoint score.

\paragraph{WeightWatch-SVD.}
Watch the Weights analyzes leading singular directions of the weight difference
between a fine-tuned model and its base model. The original method operates on
$o_{\mathrm{proj}}$ and MLP $down_{\mathrm{proj}}$ updates and combines these
directions with activation-based monitoring and clean-data calibration.
However, our canonical LoRA checkpoints modify only
$q_{\mathrm{proj}}$ and $v_{\mathrm{proj}}$, making the original module scope
uninformative in our setting. We therefore evaluate an adapted checkpoint-level,
weights-only version, denoted WeightWatch-SVD.

For each modified module, we compute the singular subspace of the effective
LoRA update without materializing the dense matrix. Writing
$\Delta W=BA$ for the stored scaled factors, we use thin QR decompositions
\[
B=Q_B R_B,\qquad A^\top=Q_A R_A,
\]
followed by an SVD of the small core matrix
\[
R_B R_A^\top = U_c \Sigma V_c^\top .
\]
The right singular vectors are then
\[
V = Q_A V_c.
\]
We retain up to the original default of 20 directions, capped by the numerical
rank of the LoRA update.

To compare two checkpoints $i$ and $j$, we use the sign-invariant projection
kernel
\[
K_m(i,j)
=
\frac{
\|V_{i,m}^{\top}V_{j,m}\|_F^2
}{
k_{ij}
},
\]
where $k_{ij}$ is the number of shared retained directions. Module-level
similarities are averaged across all audited $q/v$ modules to obtain
$K_{\mathrm{WW}}(i,j)$.

Because the original Watch the Weights method does not define a
checkpoint-level scalar score, we use the same frozen $C/S/B$ reference split
as TRACE to convert this representation into a continuous audit score. Let
$\mu_C$ be the harmful-compliance prototype and
$\mu_N=\tfrac{1}{2}(\mu_S+\mu_B)$ the balanced non-harmful prototype in the
WeightWatch-SVD kernel space. We fit
\[
x_\theta
\approx
\gamma_C\mu_C+\gamma_N\mu_N,
\qquad
\gamma_C,\gamma_N\geq0,
\]
and report
\[
s_{\mathrm{WW}}
=
\frac{\gamma_C}
{\gamma_C+\gamma_N+\epsilon}.
\]
No model queries, calibration data, ratio-specific tuning, or test labels are
used. We therefore refer to this method throughout as
\emph{WeightWatch-SVD (adapted)}, rather than as an unmodified reproduction of
the original monitoring pipeline.

\end{document}